\pdfoutput=1
\documentclass{article} %
\usepackage{placeins}
\usepackage{iclr2027_conference,times}
\def\papermode{arxiv}
\def\modesubmission{submission}\def\modecamera{camera}
\ifx\papermode\modesubmission\else\iclrfinalcopy\fi
\usepackage{sansmath}   %

\usepackage{graphicx}
\usepackage{amsmath}
\usepackage{booktabs}
\usepackage{wrapfig}
\usepackage{multirow}
\usepackage{xcolor}
\usepackage{colortbl}
\usepackage{tikz}
\usetikzlibrary{positioning,arrows.meta,calc,fit,backgrounds,shadows,shapes.geometric}
\usepackage{xspace}
\newcommand{\mname}{CADRE\xspace}   %

\usepackage[skins,breakable]{tcolorbox}
\definecolor{takeawayC}{HTML}{1F7D52}
\definecolor{takeawaybgC}{HTML}{F2FAF5}
\tcbset{takeawaystyle/.style={
    enhanced, colback=takeawaybgC, colframe=takeawayC, boxrule=1pt,
    arc=3mm, drop fuzzy shadow={takeawayC!25},
    left=8pt, right=8pt, top=11pt, bottom=4pt,
    before skip=9pt, after skip=6pt, fontupper=\small,
    coltitle=takeawayC, colbacktitle=white, fonttitle=\small\bfseries,
    attach boxed title to top left={xshift=10pt, yshift=-\tcboxedtitleheight/2},
    boxed title style={colframe=takeawayC, boxrule=0.8pt, arc=2pt,
                       left=5pt, right=5pt, top=2pt, bottom=2pt}
}}
\newtcolorbox{takeaway}[1]{takeawaystyle, unbreakable, title={#1}}
\newtcolorbox{tldrbox}[1][]{takeawaystyle, breakable, before upper={\setlength{\leftmargini}{1.2em}},
    title={Takeaway}, #1}

\definecolor{cvprblue}{rgb}{0.21,0.49,0.74}
\definecolor{darkpastelblue}{rgb}{0.47, 0.62, 0.8}
\definecolor{lightpastelblue}{rgb}{0.75, 0.85, 0.95}

\usepackage[hidelinks,colorlinks=true,linkcolor=darkpastelblue,citecolor=darkpastelblue]{hyperref}
\usepackage{url}

\title{Rethinking Video Autoencoders\\ for Interactive World Models}

\ificlrfinal
\author{{\fontsize{9.5pt}{11pt}\selectfont Adrien Ramanana Rahary$^{1,2}$\hspace{0.7em}Yutong Bai$^{3}$\hspace{0.7em}Nicolas Dufour$^{1}$\hspace{0.7em}Patrick P\'erez$^{1,*}$\hspace{0.7em}David Picard$^{2}$}\\
{\small $^{1}$Kyutai\qquad $^{2}$LIGM, ENPC, IP Paris, CNRS, UGE\qquad $^{3}$UC Berkeley}\\
{\small\texttt{\{adrien.ramanana-rahary,\,david.picard\}@enpc.fr}}\\
{\small\texttt{nicolas.dufour@kyutai.org}\quad \texttt{yutongbai@berkeley.edu}\qquad $^{*}$Work done while at Kyutai.}}
\else
\author{Anonymous authors\\Paper under double-blind review}
\fi

\begin{document}

\maketitle
\ificlrfinal\vspace{-8pt}\ifx\papermode\modecamera\else\lhead{}\fi\fi   %

\begin{abstract}
Video world models predict future frames from past frames and actions. They serve as simulators in which agents train, plan or are evaluated, and that
humans can control interactively. They generate in the latent space
of a video autoencoder, whose compression sets their cost per frame and how long each
action waits to take effect. Yet world models often reuse the autoencoders of text-to-video
generators, built for offline generation with no agent waiting and, without their training data
and code, not retrainable at another compression. We therefore ask how, where and how much a world model
should compress. We build \mname, an open family of causal diffusion representation
autoencoders spanning multiple spatio-temporal compressions. Its members differ only in a small bottleneck on
frozen features and share a diffusion decoder self-distilled to one or a few steps.
We train the same world model on each autoencoder in two
domains, manipulation and navigation, with two environments each. \mname forms most of the latency-quality frontier, especially with a moving camera. Our study also leads to four findings. At strong
compression, a diffusion decoder is sharper than perceptual regression. In robot manipulation, temporal compression is more effective in the autoencoder than in the world model. Manipulation and navigation call for opposite compressions:
temporal compression costs little when the camera is still, and spatial compression when it moves. Surprisingly,
with a still camera, coarser spatial tokens hurt motion far more than image quality.
\ificlrfinal\ifx\papermode\modecamera Code, models and a live demo are at \url{https://cadre-wm.kyutai.org}.\else Code, models and a live demo are at \url{https://kyutai.org/blog/2026-10-22-cadre}.\fi\else We will release code and models; rollouts are at \url{https://wmoptimal.github.io}.\fi
\parfillskip=0pt
\end{abstract}

\section{Introduction}
\suppressfloats[t]   %

\begin{figure}[!b]
\centering
\includegraphics[width=\textwidth]{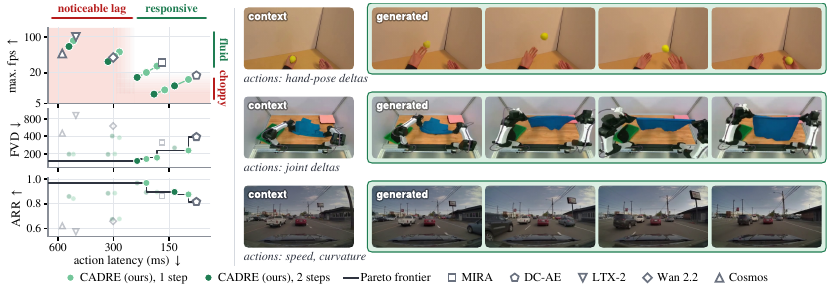}
\caption{\textbf{The right video autoencoder lets interactive world models react fast with high image and
motion quality.} On \mname, world models run at 20 fps on one H100 and react in under 230~ms, below noticeable action
lag~\citep[top left;][]{shimada2009detectionvisualfeedbackdelay}, and form most of the latency-quality frontier in
motion fidelity (FVD) and action following (ARR; bottom left). Right: rollouts given actions,
where action lag makes acting hard (juggling, folding) or dangerous (driving).}
\label{fig:teaser}
\end{figure}

A \emph{world model} is a learned simulator of an environment~\citep{ha2018worldmodels}. Given past frames and an agent's actions, a \emph{video world model} generates the next frames, and can be used to train agents~\citep{hafner2025dreamerv3,alonso2024diamond}, evaluate robot policies~\citep{yang2024unisim}, plan the outcome of actions~\citep{bar2025navigationworldmodels}, or simply play a game~\citep{bruce2024genie,valevski2024gamengen,hu2026multiplayerinteractiveworldmodels}.
To enable all these applications, two different aspects have to be optimized: the \emph{throughput} that measures the number of frames generated per second, and the \emph{latency} that measures the delay between the action and its effect on the output.
Throughput limits how fast a simulation can happen, whereas latency limits how reactive and faithful a simulation can be.

Most video world models operate in the latent space of a video autoencoder~\citep{bruce2024genie,valevski2024gamengen}.
To avoid designing an \textit{ad hoc} autoencoder, many methods~\citep{he2025matrixgame2,li2025hunyuangamecraft,zhu2026sanawmefficientminutescaleworld} reuse the autoencoder of a text-to-video generator such as LTX-2~\citep{hacohen2026ltx2efficientjointaudiovisual}, Wan~2.2~\citep{wan2025openadvancedlargescalevideo} or HunyuanVideo~\citep{kong2024hunyuanvideo}. 
They thus inherit the throughput and latency that come with it.
Yet, such autoencoders were not designed with the throughput-latency trade-off of world models in mind, and often perform massive temporal compression since it has low impact on visual quality.
A typical $8{\times}$ temporal compression delays every action by at least 400~ms at 20 frames per second. Humans
notice when their own movement appears more than about 230~ms late~\citep{shimada2009detectionvisualfeedbackdelay},
and interaction feels immediate only below 100~ms~\citep{miller1968responsetime,claypool2006latencyplayeractions}. 
This raises the question of how other throughput-latency trade-offs affect rollout quality and action following.

To address this question, we propose \mname,
an open family of \textbf{ca}usal \textbf{d}iffusion \textbf{re}presentation autoencoders built for interactive world models. Its members differ only in a small bottleneck on frozen features, which sets the compression,
and share a causal diffusion decoder that needs one or a few steps. To study compression across settings, we train the same world model, a causal diffusion transformer following
MIRA~\citep{hu2026multiplayerinteractiveworldmodels}, on each autoencoder, in two domains: manipulation (by a robot and by human hands) and navigation (driving and
indoor). We compare \mname with the autoencoders of Wan~2.2, Cosmos~\citep{cosmos2025worldfoundationmodelplatform}
and LTX-2, the image autoencoder DC-AE~\citep{chen2025deepcompressionautoencoderefficient} and the MIRA autoencoder
retrained on our data. \mname forms most of the latency-quality frontier, especially with a moving camera (Figure~\ref{fig:teaser}).

Interestingly, our study points to four findings. (1)~At strong compression, a diffusion decoder is sharper than perceptual regression, and a few-step flow model trained by progressive self-distillation is a good fit. (2)~In robot manipulation, temporal compression is more effective in the
autoencoder than in the world model: patching inside the world model loses detail,
and at eight frames also action following. (3)~Manipulation and navigation call for opposite compressions. In manipulation the camera is still, so
temporal compression costs little while spatial compression is costly. In navigation the camera moves and the balance
reverses, so it pays to keep frames and compress spatially, the low-latency choice. (4)~Surprisingly, with a still camera, coarser spatial tokens hurt motion far more than image quality, and, in robot manipulation, action following too.

\textbf{Contributions.} \begin{itemize}
  \item \textbf{What compression controls in an interactive world model.} We make explicit how an autoencoder's spatial and temporal compression set a world model's
  throughput and action latency, and we separate how much to compress from where to
  compress (Section~\ref{sec:controls}).
  \item \textbf{\mname, causal diffusion representation autoencoders for interactive world models.} Open models at seven
  spatio-temporal compressions, whose decoder learns one- or few-step decoding by progressive self-distillation, in a single stage and without an
  adversarial loss (Section~\ref{sec:method}).
  \item \textbf{Where, and how much, to compress video for a world model.} A controlled study in four environments, judged by rollouts and action
following (Sections~\ref{sec:family} and~\ref{sec:compress}).
\end{itemize}

\section{What Compression Controls in an Interactive World Model}
\label{sec:controls}

Before studying how to compress, we make explicit what compression controls. An interactive world model generates
in the latent space of an autoencoder that compresses a clip by $\tau$ temporally and $s$ spatially, written $\tau{\times}s{\times}s$: $2{\times}32{\times}32$ compresses $2\times$ temporally
and $32\times$ spatially. These two factors set the world model's cost, action latency and throughput. The
compression itself can happen in the autoencoder or in the world model (Section~\ref{sec:placement}).

\textbf{Compression sets the world model's cost.} The world model we study is a causal, space-time-factorized diffusion transformer that predicts the next latent
frame from the previous ones and the current action or action chunk (Section~\ref{sec:experiments}, Appendix~\ref{app:wm}). Its cost grows with the number
of tokens it processes, which only the spatial and temporal compression set: a clip of $T$ frames of size $H{\times}W$
becomes $N = (T/\tau)(H/s)(W/s)$ latent tokens for the world model.

\begin{figure}[t]
\centering
\input{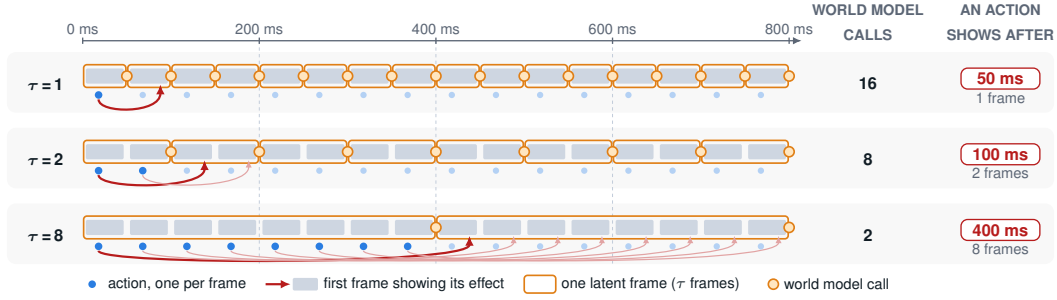}
\resizebox{\textwidth}{!}{\begin{tikzpicture}[x=1cm, y=1cm, font=\sffamily\sansmath,
  shadowed/.style={drop shadow={shadow xshift=0pt, shadow yshift=-0.7pt, opacity=0.14, fill=inkC}},
]
\def\fA{\fontsize{7.4}{8.6}\selectfont}                %
\def\fB{\fontsize{6.6}{7.8}\selectfont}                %
\def\W{11.2}\pgfmathsetmacro{\sw}{\W/16}               %
\def\rh{1.2}                                          %
\def\xS{\W+1.25}\def\xL{\W+3.35}                        %
\pgfmathsetmacro{\yT}{0.62}
\pgfmathsetmacro{\yBot}{-2*\rh-0.62}
\foreach \r in {0,...,2}{
  \fill[inkC!3, rounded corners=4pt] (-1.2,-\r*\rh-0.52) rectangle (\xL+0.85,-\r*\rh+0.44);
}
\foreach \k in {4,8,12}{
  \draw[lineC, line width=0.4pt, dash pattern=on 1.5pt off 1.5pt] (\k*\sw,\yT) -- (\k*\sw,\yBot+0.1);
}
\draw[mutedC, line width=0.6pt, -{Latex[length=1.4mm,width=1.2mm]}] (0,\yT) -- (\W+0.2,\yT);
\foreach \k/\ms in {0/0,4/200,8/400,12/600,16/800}{
  \draw[mutedC, line width=0.6pt] (\k*\sw,\yT-0.05) -- (\k*\sw,\yT+0.05);
  \node[anchor=south, text=mutedC, inner sep=1.5pt] at (\k*\sw,\yT+0.05) {\fB\ms~ms};
}
\node[anchor=south, text=mutedC, inner sep=1.5pt, align=center] at (\xS,\yT-0.05) {\fB\bfseries WORLD MODEL\\\fB\bfseries CALLS};
\node[anchor=south, text=mutedC, inner sep=1.5pt, align=center] at (\xL,\yT-0.05) {\fB\bfseries AN ACTION\\\fB\bfseries SHOWS AFTER};
\foreach \t/\r/\ms in {1/0/50,2/1/100,8/2/400}{
  \pgfmathsetmacro{\yr}{-\r*\rh+0.06}\pgfmathsetmacro{\ya}{\yr-0.3}
  \pgfmathtruncatemacro{\nc}{16/\t}\pgfmathtruncatemacro{\tl}{\t-1}
  \node[anchor=east, text=inkC, inner sep=0pt] at (-0.3,\yr-0.12) {\fA\bfseries $\boldsymbol{\tau}$\,=\,\t};
  \foreach \k in {0,...,15}{
    \fill[lineC!80, rounded corners=0.8pt] (\k*\sw+0.05,\yr-0.12) rectangle (\k*\sw+\sw-0.05,\yr+0.12);
  }
  \foreach \j in {1,...,\nc}{
    \draw[trnbdC, line width=0.75pt, rounded corners=1.5pt] ({(\j-1)*\t*\sw+0.015},\yr-0.18) rectangle ({\j*\t*\sw-0.015},\yr+0.18);
  }
  \foreach \j in {1,...,\nc}{
    \fill[trnflC, draw=trnbdC, line width=0.75pt] ({\j*\t*\sw},\yr) circle (0.08);
  }
  \foreach \i in {0,...,\tl}{
    \pgfmathsetmacro{\xa}{(\i+0.36)*\sw}\pgfmathsetmacro{\xb}{(\i+\t+0.76)*\sw}
    \ifnum\i=0
      \draw[hiC, line width=1.0pt, -{Latex[length=1.6mm,width=1.4mm]}]
        (\xa,\ya-0.07) .. controls (\xa,\ya-0.3) and (\xb,\ya-0.3) .. (\xb,\yr-0.13);
    \else
      \draw[hiC!40, line width=0.55pt, -{Latex[length=1.2mm,width=1.0mm]}]
        (\xa,\ya-0.07) .. controls (\xa,\ya-0.3) and (\xb,\ya-0.3) .. (\xb,\yr-0.13);
    \fi
  }
  \foreach \k in {0,...,15}{
    \ifnum\k<\t
      \fill[actbdC] ({(\k+0.36)*\sw},\ya) circle (0.06);
    \else
      \fill[actbdC!35] ({(\k+0.36)*\sw},\ya) circle (0.05);
    \fi
  }
  \node[text=inkC, inner sep=0pt] at (\xS,\yr-0.12) {\fA\bfseries\nc};
  \node[fill=white, draw=hiC, line width=0.6pt, rounded corners=3pt, text=hiC, inner xsep=5pt, inner ysep=2pt,
        minimum width=1.25cm] at (\xL,\yr-0.04) {\fA\bfseries\ms~ms};
  \node[text=mutedC, inner sep=0pt] at (\xL,\yr-0.34) {\fB\t{} frame\ifnum\t>1 s\fi};
}
\pgfmathsetmacro{\yK}{\yBot-0.22}
\fill[actbdC] (0.15,\yK) circle (0.06);
\node[anchor=west, text=inkC, inner sep=1pt] at (0.36,\yK) {\fB action, one per frame};
\draw[hiC, line width=1.0pt, -{Latex[length=1.6mm,width=1.4mm]}] (2.9,\yK) -- (3.3,\yK); \fill[lineC!80, rounded corners=0.8pt] (3.33,\yK-0.1) rectangle (3.73,\yK+0.1);
\node[anchor=west, text=inkC, inner sep=1pt] at (3.78,\yK) {\fB first frame showing its effect};
\draw[trnbdC, line width=0.75pt, rounded corners=1.5pt] (7.0,\yK-0.14) rectangle (7.5,\yK+0.14);
\node[anchor=west, text=inkC, inner sep=1pt] at (7.55,\yK) {\fB one latent frame ($\tau$ frames)};
\fill[trnflC, draw=trnbdC, line width=0.75pt] (10.95,\yK) circle (0.08);
\node[anchor=west, text=inkC, inner sep=1pt] at (11.08,\yK) {\fB world model call};
\end{tikzpicture}}
\caption{\textbf{Every frame registers an action, whose effect appears $\tau$ frames later.} A latent frame is predicted
only once all its $\tau$ actions are in, so a larger $\tau$ saves world model calls but delays actions.}
\label{fig:chunk}
\end{figure}

\textbf{A latent frame waits for all its actions.} Every frame registers one action. A latent frame spans $\tau$ frames, so the world model can predict it only once
all $\tau$ conditioning actions are known. It therefore consumes actions in chunks, and the actions of one chunk shape the next
(Figure~\ref{fig:chunk}). The agent thus commits to $\tau$ actions before seeing their effect, as action-chunking policies do~\citep{zhao2023learningfinegrainedbimanualmanipulation,chi2024diffusionpolicyvisuomotorpolicy,black2026pi0visionlanguageactionflowmodel}. Causality holds between chunks. Within a chunk, the $\tau$ frames
decode from the same latent, and their order is learned from real video.

\textbf{Temporal compression delays every action.} The first action of a chunk waits $\tau/f$ for the chunk to close, where $f$ is the
frame rate, then $t_{\text{gen}}$ for the world model to generate the latent frame and $t_{\text{dec}}$ for the autoencoder to decode
it into $\tau$ frames, which sets the throughput:
\begin{equation}
\text{latency} = \underbrace{\tau/f}_{\text{chunk closes}} + \underbrace{t_{\text{gen}} + t_{\text{dec}}}_{\text{generate and decode}},
\qquad
\text{throughput} = \frac{\tau}{t_{\text{gen}} + t_{\text{dec}}}.
\label{eq:latency}
\end{equation}
No speed-up shortens the first term, which at $\tau=8$ and 20 frames per second is 400~ms, beyond the 230~ms lag humans
notice in their own movement~\citep{shimada2009detectionvisualfeedbackdelay}. In exchange, a larger $\tau$ raises the
throughput and, at a fixed spatial factor, cuts the token count $N$.

\section{\mname: Causal Diffusion Representation Autoencoders}
\label{sec:method}

\begin{figure}[t]
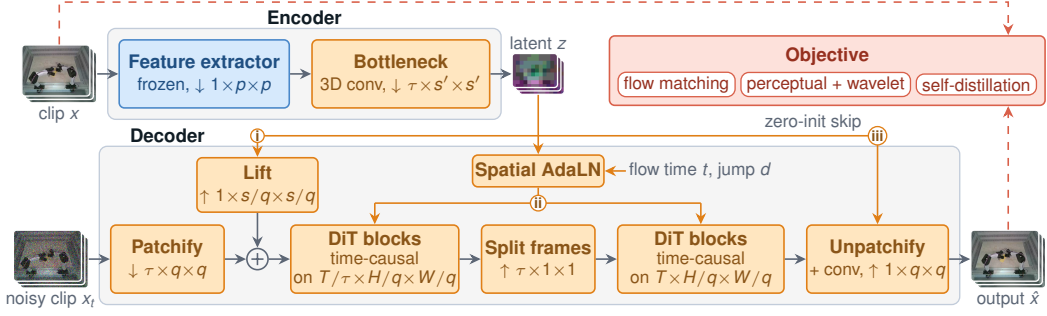

\centering
\input{figures/codec_defs}
\input{figures/codec_v2}
\caption{\textbf{\mname.} A frozen feature extractor (blue; DINOv3) and a
trained bottleneck compress the clip by $\tau{\times}s{\times}s$. A causal diffusion decoder denoises it, conditioned
on the latent in three places (i--iii, Section~\ref{sec:decoder}). Only the bottleneck, lift and temporal patch size depend on $\tau{\times}s{\times}s$.}
\label{fig:arch}
\end{figure}

Studying compression means training the autoencoder at many compression points. The autoencoders of
text-to-video generators release their weights but not their training data or code, so they cannot be retrained. We instead design an open family of autoencoders for world models, one per
compression point (Figure~\ref{fig:arch}). Its members share one recipe, differ only in their
compression and are each trained from scratch, so the world models trained on them differ by compression alone.

\subsection{Architecture}
\label{sec:bottleneck}
\label{sec:decoder}

\textbf{Frozen features and one trained layer encode the clip.} As in MIRA~\citep{hu2026multiplayerinteractiveworldmodels}, the encoder is that of a representation
autoencoder~\citep{zheng2025diffusiontransformersrepresentationautoencoders}. A frozen DINOv3
extractor~\citep{simeoni2025dinov3} encodes each frame into one feature per $p{\times}p$ patch, aggregating several
layers as in RAEv2~\citep{singh2026improvedbaselinesrepresentationautoencoders}. One trained 3D convolution with
kernel $(\tau, k, k)$ and stride $(\tau, s', s')$ compresses $\tau$ frames of $s'^2$ patches into a deterministic
latent, with $s = p\,s'$ (Appendix~\ref{app:ae}).

\textbf{A causal diffusion decoder keeps textures sharp.} A strongly compressed latent keeps a scene's layout and semantics but little texture, so a perceptual regression decoder
averages the textures consistent with it into a blur~\citep{xu2026pointditpixelspacediffusionmonocular}. We therefore use a diffusion
decoder~\citep{preechakul2022diffusionautoencodersmeaningfuldecodable,pernias2023wuerstchenefficientarchitecturelargescale,chen2025diffusionautoencodersscalableimage},
which samples the clip conditioned on the latent. It is a space-time-factorized
transformer~\citep{bertasius2021spacetimeattentionneedvideo,arnab2021vivitvideovisiontransformer} on a $q{\times}q$
patch grid, whose blocks attend over space, then causally over time. Most blocks process one token per latent frame, that is,
per $\tau$ video frames. A linear map then splits each token into $\tau$ tokens, one per frame, so the last blocks
resolve motion within a latent frame. The latent conditions the diffusion decoder in three places: (i)~lifted by a transposed convolution and summed with the patchified noisy clip, (ii)~through
a spatial AdaLN~\citep{park2019spade,zhu2020sean,dufour2022scam,yu2026pixelditpixeldiffusiontransformers} in every block, which also receives the flow time
$t$ and the jump length $d$, the size of the denoising step the decoder takes (Section~\ref{sec:objective}, Appendix~\ref{app:adaln}), and (iii)~through a zero-initialized skip into a light
convolutional head. The decoder is time-causal, so it streams chunk by chunk.

\subsection{Training}
\label{sec:objective}

\textbf{The decoder learns by flow matching and perceptual losses.} The decoder $g_\theta$ uses the $x$-prediction, $x$-loss form of flow
matching~\citep{li2026basicsletdenoisinggenerative}. It predicts the clean clip $x$ from the latent $z$ and a
noisy clip $x_t = t\,x + (1-t)\,\epsilon$, with $\epsilon\sim\mathcal{N}(0,I)$ and $t=0$ pure noise. As in diffusion
forcing~\citep{chen2024diffusionforcingnexttokenprediction}, each chunk of $\tau$ frames draws its own
$t\sim\mathcal{U}(0,1)$. The loss is
\begin{equation}
\mathcal{L}_{\text{fm}} \;=\; \mathrm{E}_{x,\epsilon,t}\;
\ell_{\text{rec}}\big(g_\theta(z,\, x_t,\, t,\, 0),\; x\big),
\label{eq:flow}
\end{equation}
where the last argument, the jump length $d$, is zero here. The term $\ell_{\text{rec}}$ sums a mean absolute error,
two perceptual distances (LPIPS~\citep{zhang2018unreasonableeffectivenessdeepfeatures} and its analog in DINO
feature space) and a 3D wavelet term as in LTX-Video~\citep{hacohen2024ltxvideorealtimevideolatent}, weighted by
gradient balancing. \mname trains on a 76k-hour mixture of web, robot, driving and camera video (Appendix~\ref{app:data}).

\textbf{Self-distillation makes one or two decoding steps enough.} Flow-matching trajectories curve, so following them takes many steps, while an interactive world model needs single-
or few-step decoding. Shortcut models~\citep{frans2025stepdiffusionshortcutmodels} and self-distilled flow
maps~\citep{boffi2025buildconsistencymodellearning} condition the network on its step size and train one large
step to agree with two smaller ones. Following them, $g_\theta(z, x_r, r, d)$ predicts where a jump from $r$ to $r{+}d$
lands, through $x_{r+d} = x_r + \tfrac{d}{1-r}(\hat{x}-x_r)$ with implied velocity $v_\theta = (\hat{x}-x_r)/(1-r)$. One
jump from $r$ to $t$ is trained to land where two half jumps through $u=(r+t)/2$ do (dropping $z$ from the notation),
\begin{equation}
\mathcal{L}_{\text{psd}} \;=\; \mathrm{E}\,\Big\|(1-r)\Big(v_\theta(x_r, r, t{-}r)
- \operatorname{sg}\big[\tfrac{1}{2}\big(v_\theta(x_r, r, u{-}r) + v_\theta(\hat{x}_u, u, t{-}u)\big)\big]\Big)\Big\|_2^2,
\label{eq:psd}
\end{equation}
where $\operatorname{sg}$ stops the gradient and $\hat{x}_u$ is the point reached by the first half jump. Unlike the
consistency decoder~\citep{openai2023consistencydecoder} and SSDD~\citep{vallaeys2026ssddsinglestepdiffusiondecoder},
which distill a decoder in a second stage, the \mname decoder is its own teacher and learns one- or few-jump decoding in one stage.

\section{Judging an Autoencoder by Its World Model}
\label{sec:experiments}

An autoencoder is trained to reconstruct its input, yet in latent generation, reconstruction and generation
trade off~\citep{yao2025reconstructionvsgenerationtaming,hansenestruch2025learningsscalingvisualtokenizers}, and a
line of work shapes the latent for the generator
instead~\citep{leng2025repaeunlockingvaeendtoend,kouzelis2025eqvaeequivarianceregularizedlatent,zheng2025diffusiontransformersrepresentationautoencoders,xie2026videoraetamingvideofoundation,ramananarahary2026diffusablelatents}.
We therefore score rollouts instead of reconstructions (Appendix~\ref{app:recon}). As in MIRA~\citep{hu2026multiplayerinteractiveworldmodels},
we judge each autoencoder by the rollouts of a world model trained on its latents; the world model, clips
and actions are shared, so only the latent varies.

\textbf{Four environments span manipulation and navigation.} We train world models in four environments, two of manipulation and two of navigation (Table~\ref{tab:envs}), so
that no conclusion rests on a single environment. In manipulation the view is static for the robot and nearly static for the head-mounted camera of the
human, so mainly the arms or hands and the objects they handle move. In navigation the view itself moves, so the whole frame changes; we
separate driving, a dynamic scene where other road users move on their own, from indoor navigation in static scenes.
Action complexity ranges from 2 dimensions for driving to 102 for two hands. All videos are resampled to 20 frames per second at
$288{\times}512$. Evaluation uses held-out episodes, and the official split for driving.

\begin{table}[t]
\centering\small
\caption{\textbf{Environments.} One world model is trained per environment and per autoencoder.}
\label{tab:envs}
\setlength{\tabcolsep}{3pt}
\begin{tabular}{lllllr}
\toprule
& Environment & Dataset & Camera & Action & DoF \\
\midrule
\multirow{2}{*}{Manipulation} & Robot & ABC~\citep{allshire2026abc} & static & joint deltas & 14 \\
& Human & EgoDex~\citep{hoque2026egodexlearningdexterousmanipulation} & nearly static & hand-pose deltas & 102 \\
\midrule
\multirow{2}{*}{Navigation} & Driving & Physical AI AV~\citep{nvidia2025physicalaiav} & moving & speed, curvature & 2 \\
& Indoor & RealEstate10K~\citep{zhou2018stereo} & moving & camera twist & 6 \\
\bottomrule
\end{tabular}
\end{table}

\textbf{One world model recipe serves every autoencoder.} The world model is a 1.2B-parameter causal diffusion
transformer conditioned on the $\tau$ actions of each latent frame and trained with diffusion
forcing~\citep{chen2024diffusionforcingnexttokenprediction} (Appendix~\ref{app:wm}). Every autoencoder,
\mname included, uses the same recipe from MIRA~\citep{hu2026multiplayerinteractiveworldmodels}, with its latents
standardized by their own statistics. Following SD3~\citep{esser2024scalingrectifiedflowtransformers} and
RAE~\citep{zheng2025diffusiontransformersrepresentationautoencoders}, our tuning includes shifting the noise schedule with
each autoencoder's latent size and token count.

\textbf{Rollouts are scored on image and motion quality, and on action following.} Each world model receives 40 real frames (2~s) with their actions and generates the next 96 (4.8~s) under the
recorded actions, on 2048 held-out samples per environment shared across autoencoders. We score the last 16
decoded frames. FID~\citep{heusel2018ganstrainedtimescaleupdate} measures image quality and
FVD~\citep{unterthiner2019accurategenerativemodelsvideo} also penalizes implausible motion and temporal seams. PSNR and
LPIPS~\citep{zhang2018unreasonableeffectivenessdeepfeatures} measure closeness to the recorded future; PSNR should always be accompanied by distributional metrics, as it rewards
blur. Action following is measured by the action recoverability ratio (ARR), which extends the inverse dynamics evaluation of
MIRA~\citep{hu2026multiplayerinteractiveworldmodels}: a frozen probe reads the actions back from the rollout, and
their coefficient of determination ($R^2$) is normalized by the probe's $R^2$ on real video, so an ARR of one means real-video recoverability
(Appendix~\ref{app:idm}).

\textbf{We compare seven \mname autoencoders with five recent ones.}\label{sec:family}\label{sec:sotaSec} The seven differ only in their compression $\tau{\times}s{\times}s$, and we name each \mname-$\tau{\times}s{\times}s$ (Figure~\ref{fig:members}, Appendix~\ref{app:full}). Of the others, four are off the shelf and used at their native compression:
Wan~2.2~\citep{wan2025openadvancedlargescalevideo} ($4{\times}16{\times}16$),
Cosmos~\citep{cosmos2025worldfoundationmodelplatform} ($8{\times}8{\times}8$),
LTX-2~\citep{hacohen2026ltx2efficientjointaudiovisual} ($8{\times}32{\times}32$) and the image
autoencoder DC-AE~\citep{chen2025deepcompressionautoencoderefficient} ($1{\times}32{\times}32$). The fifth retrains the MIRA recipe~\citep{hu2026multiplayerinteractiveworldmodels} on our data at
$2{\times}32{\times}32$. The LTX-2 decoder reads future latents, so we decode each prefix and keep its newest
frames, as a streaming system would. The generators they come from train on far more video, such as 20M curated hours for Cosmos, likely
including many datasets used here.

\begin{figure}[t]
\centering
\begin{minipage}[t]{0.485\textwidth}\centering
\textbf{(a)}\\[-1pt]\includegraphics[width=\textwidth]{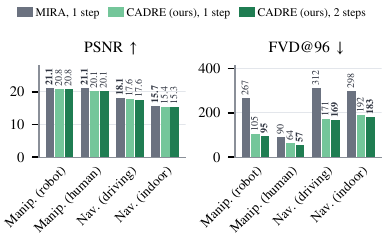}
\end{minipage}\hfill
\begin{minipage}[t]{0.485\textwidth}\centering
\textbf{(b)}\\[-1pt]\includegraphics[width=\textwidth]{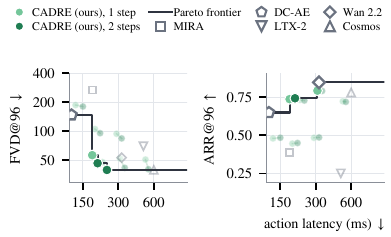}
\end{minipage}
\caption{\textbf{(a) Diffusion against perceptual regression decoding} at $2{\times}32{\times}32$: MIRA has the worst FVD
but the best PSNR, a sign of blur (all metrics: Figure~\ref{fig:decoderfull}). \textbf{(b) Rollout quality
against action latency} in robot manipulation: \mname forms most of the frontier (all environments:
Figure~\ref{fig:speedapp}).}
\label{fig:decspeed}
\end{figure}

\subsection{Should the decoder use perceptual regression or diffusion?}
\label{sec:genDecoder}
A decoder can regress one video from the latent or sample one by diffusion. MIRA~\citep{hu2026multiplayerinteractiveworldmodels}
takes the first route: it replaces the adversarial losses used since VQGAN~\citep{esser2021tamingtransformershighresolutionimage}
with perceptual losses (LPIPS and DINO) and works well on Rocket League. It is also the only autoencoder besides \mname
that runs at 20 frames per second within the 230~ms lag humans notice (Figure~\ref{fig:teaser}). To test this route beyond
games, we retrain the MIRA recipe on our mixture (Appendix~\ref{app:data}) at $2{\times}32{\times}32$ with
our extractor and budget (Figure~\ref{fig:decspeed}a). Against the \mname decoder, it gives higher PSNR but worse FID and FVD in every environment. In one step, the \mname decoder lowers FVD by 1.4 to 2.5$\times$ with similar
action following. A second step lowers FID by up to 23\%.

\begin{takeaway}{Should the decoder use perceptual regression or diffusion?}
\textbf{Decode a compressed latent by diffusion.} Such a latent fits many videos. Perceptual regression outputs
their blurry average, while diffusion outputs one sharp video. Self-distilled, it decodes in one step as fast as
regression with 1.4 to 2.5$\times$ lower FVD and similar action following; a second step improves it further.
\end{takeaway}

\subsection{How do the autoencoders trade quality for latency?}
\label{sec:quality}
To compare the autoencoders on both axes at once, we plot rollout quality against action latency
(Figure~\ref{fig:decspeed}b for robot manipulation, Figure~\ref{fig:teaser} for driving, Figure~\ref{fig:speedapp} for all); an autoencoder lies on the Pareto frontier when no other is both
faster and better. \mname autoencoders form most of this frontier in every environment. In 42 of the 48 comparisons
with MIRA, Wan~2.2, Cosmos and LTX-2 (four environments; FID, FVD and ARR), a \mname autoencoder is at least as fast
and better by more than 2\%; \mbox{DC-AE}, faster than all, is compared below. With two decoding
steps, the best FID of every environment belongs to \mname, most often to \mname-$4{\times}16{\times}16$, and so does the
best FVD, except in robot manipulation, where \mname and Cosmos tie. The gap widens with a moving camera: in navigation, all \mname autoencoders but one beat
every other on FID and FVD, while Wan~2.2's FVD grows from 53 in robot manipulation to 600 or more in navigation and LTX-2's ARR falls
to 0.04 indoors. \mname also drifts least in 64~s robot manipulation rollouts (Appendix~\ref{app:drift}).

\begin{figure}[t]
\centering
\includegraphics[width=\textwidth]{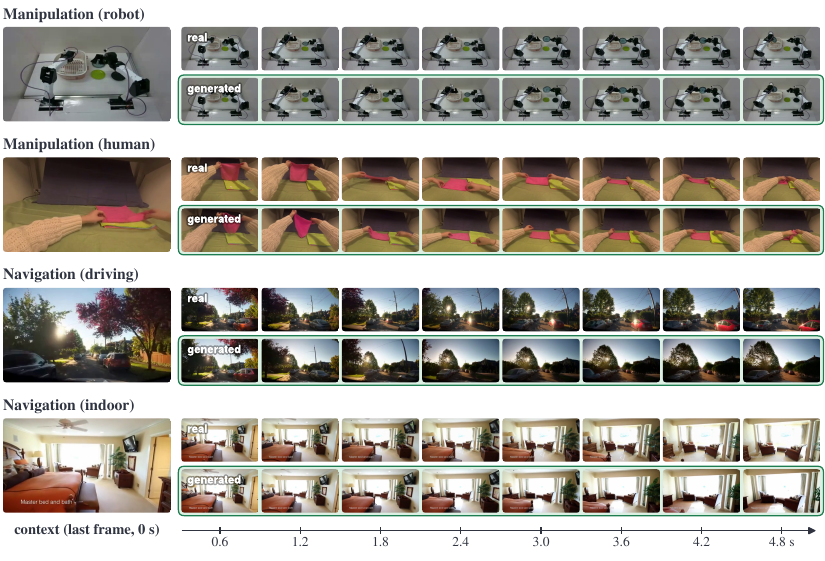}
\caption{\textbf{Real-time rollouts following recorded actions.} From 2~s of real video (left) and its actions, \mname world
models generate all other frames, with contact in robot manipulation, cloth dynamics in human manipulation,
plausible futures in driving and 3D-consistent indoor scenes.}
\label{fig:qualitative}
\end{figure}

At the fast end, \mname-$1{\times}48{\times}48$ reacts about as fast as \mbox{DC-AE} and beats it in navigation; \mbox{DC-AE} wins in manipulation. Few autoencoders are both fluid and responsive: streaming on one H100 with eight denoising steps, only
\mname-$2{\times}32{\times}32$, \mname-$2{\times}16{\times}16$ and MIRA run at 20 frames per second under the 230~ms lag humans
notice~\citep{shimada2009detectionvisualfeedbackdelay} (Figure~\ref{fig:teaser}). The world models of
Figure~\ref{fig:qualitative} run fluidly, at 29 to 58 frames per second (Appendix~\ref{app:latency}), and,
from the same context, follow actions never recorded with it, such as holding
still or turning the other way (Figure~\ref{fig:controllability}).

\section{How to Compress for Interactive World Models}
\label{sec:compress}

We now ask three questions: which component should compress temporally, how many frames one latent
should span, and what coarser spatial tokens trade for their speed. Comparisons follow the protocol of
Section~\ref{sec:experiments}, so arms differ only in how the video is compressed.

\subsection{Should the autoencoder or the world model compress temporally?}
\label{sec:placement}

\begin{figure}[t]
\centering
\includegraphics[width=\textwidth]{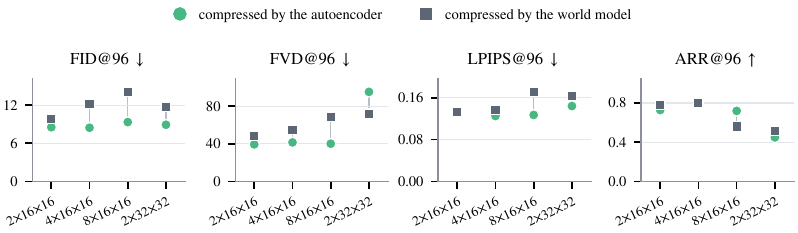}
\caption{\textbf{Temporal compression in the autoencoder or in the world model}, at the same token grid in robot
manipulation. The autoencoder gives sharper frames, closer to the recorded future, and more plausible motion at all but
$2{\times}32{\times}32$. Action following barely changes, except at 8 frames.}
\label{fig:placement}
\end{figure}

Temporal compression can happen in the autoencoder or in the world model. The alternative to \mname is an image autoencoder that leaves frames
temporally uncompressed, with the world model patching its latent temporally. Both
routes give the same token grid, but a different model learns the compression. In robot manipulation, we reach
each grid a second time by patching the latent of \mname-$1{\times}16{\times}16$ inside the world model
(Figure~\ref{fig:placement}). Only the compression changes. Actions and noise levels stay per frame, and a patched token
receives the actions of every frame it covers.

\textbf{Temporal compression in the autoencoder gives better frames than in the world model.} Compressing in the autoencoder gives better FID, FVD, PSNR and LPIPS at every temporal grid, except FVD at $2{\times}32{\times}32$, where space is compressed too. The FVD
gap grows with the factor, from 9 points when compressing 2 frames to 28 when compressing 8. Decoder capacity does not
explain it, since the patched arm decodes with \mname-$1{\times}16{\times}16$, the best reconstruction among them
(Table~\ref{tab:recon}). We attribute this to the training objectives. The autoencoder is trained to reconstruct pixels, so its compression keeps appearance. The world model is trained to
predict latents, so its compression keeps only what helps that prediction. This matches image and video generation,
where compressing in the autoencoder beats patching~\citep{chen2025deepcompressionautoencoderefficient,gupta2023photorealisticvideogenerationdiffusion}.

\textbf{Where temporal compression happens barely changes action following.} The exception is 8 frames, where patching drops
ARR from 0.72 to 0.56. Action following depends on the motion the actions cause, which the world
model's compression is trained to keep.

\begin{takeaway}{Should the autoencoder or the world model compress temporally?}
\textbf{Let the autoencoder compress temporally.} At the same number of tokens, its compressed frames look sharper and
move more realistically, with similar action following. It learns its compression to reconstruct pixels, while the world
model keeps only what helps it predict.
\end{takeaway}

\subsection{How many frames should one latent span?}
\label{sec:tauTradeoff}
\label{sec:frontier}

\begin{figure}[t]
\centering
\includegraphics[width=\textwidth]{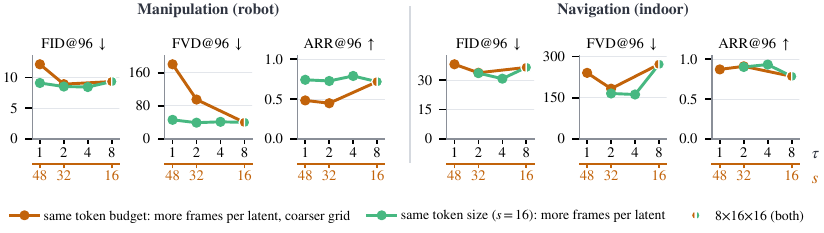}
\caption{\textbf{Temporal compression and spatial tokens.} Green raises $\tau$ at a fixed token size; orange
at a fixed token budget, so its grid coarsens (spatial factor $s$). A static view fits eight frames in one latent
at no cost; a moving view loses quality with more frames. Full results in Figure~\ref{fig:familiesfull}.}
\label{fig:families}
\end{figure}

To separate temporal compression from the cost of coarser tokens, we vary $\tau$ in two sweeps (Figure~\ref{fig:families}). The first holds $s=16$ and isolates temporal compression.
The second holds about 1100 tokens per 16-frame clip, so at a given $\tau$ its gap to the first measures the cost of coarser spatial
tokens. All world models are matched in training steps and total training frames.

\textbf{Latents should span many redundant frames in manipulation and few in navigation.} In manipulation,
the camera is mostly still, so consecutive frames are redundant and one latent can span many of them at no cost
(green curve in Figure~\ref{fig:families}). In navigation, the camera moves and the view shifts by more than a token
within one long latent (Appendix~\ref{app:motion}), so long latents degrade motion, raising FVD by up to 64\% over two-frame latents. Action
following changes less (Appendix~\ref{app:full}). Camera motion is our leading explanation, though the domains
also differ in actions and scenes.

In 64~s robot manipulation rollouts, long latents do not make errors accumulate: \mname-$8{\times}16{\times}16$ is
best from about 13~s on and ends with the lowest FVD (Appendix~\ref{app:drift}). Long latents do leave seams, visible jumps
between chunks. A boundary excess compares the frame-to-frame
change across chunk boundaries with the change within chunks, relative to real video (Appendix~\ref{app:seam}). It
grows with $\tau$ in all four environments, to 1.6--1.8 at $\tau=8$ (Table~\ref{tab:seam}). With a moving view, the seams
also show in FVD, which grows 1.6 to 2.1$\times$ more at $\tau=8$ than at $\tau=2$.

\textbf{At a fixed token budget, navigation favors few frames per latent and manipulation many.} Along the
orange curve of Figure~\ref{fig:families}, autoencoders give about 1100 tokens per 16-frame clip. In navigation, the low-latency
\mname-$2{\times}32{\times}32$ is best, with two frames and coarser tokens. In manipulation,
\mname-$8{\times}16{\times}16$ gives the most plausible motion, with eight frames and fine tokens.

\begin{takeaway}{How many frames should one latent span?}
\textbf{Manipulation and navigation call for opposite compressions.} In manipulation the camera is still, so
each token sees temporally redundant content: temporal compression costs little, but spatial compression loses
the small objects being handled. In navigation the camera moves, so each token sees new content: temporal compression
shows seams, but spatial compression costs little because the whole view shifts.
\end{takeaway}

\subsection{What do coarser spatial tokens trade for their speed?}
\label{sec:space}

At a given $\tau$, coarser spatial tokens cut the world model's cost, and the gap between the two sweeps isolates what they lose in quality and action following.

\textbf{Fine spatial tokens matter most when the action is small and stays in view.} In robot manipulation,
the action shows in a small region that stays in view. At $\tau=2$, coarser spatial tokens multiply FVD by 2.4 and
lower action following, with almost no change in FID (orange against green in Figure~\ref{fig:families}). In
navigation, the action moves the whole frame, which coarse tokens already capture, and the gap nearly vanishes
(Appendix~\ref{app:full}). FID may miss errors confined to small regions, but ARR, which reads the motion itself, drops
too (Appendix~\ref{app:full}).

\begin{takeaway}{What do coarser spatial tokens trade for their speed?}
\textbf{Surprisingly, coarser tokens hurt motion and action following far more than image quality, above all in manipulation.}
In manipulation, per-frame quality barely changes, but motion degrades, and so does action following in robot manipulation. In navigation, where the
whole view shifts, they cost little.
\end{takeaway}

\section{Related Work}

\textbf{World models and their autoencoders.} Video world models serve
driving~\citep{russell2025gaia2controllablemultiviewgenerative}, robotics~\citep{yang2024unisim} and open
scenes~\citep{bruce2024genie}. Interactive ones stream in real
time~\citep{valevski2024gamengen,alonso2024diamond}, often by distilling a text-to-video generator whose
autoencoder they inherit~\citep{he2025matrixgame2,li2025hunyuangamecraft}. Few study that autoencoder.
SANA-WM~\citep{zhu2026sanawmefficientminutescaleworld} swaps two of them offline, and
MIRA~\citep{hu2026multiplayerinteractiveworldmodels} judges its own through the world model, but only on a single game.
We compare twelve autoencoders of varied compression through world models in four environments.

\textbf{Latent structure and compression.} Generation quality depends on the structure of the
latent~\citep{rombach2022highresolutionimagesynthesislatent,yao2025reconstructionvsgenerationtaming}.
Representation autoencoders generate in the space of a frozen self-supervised
encoder~\citep{zheng2025diffusiontransformersrepresentationautoencoders,xie2026videoraetamingvideofoundation},
at one compression rate. Video autoencoders mostly compress $8{\times}$ in space and $4{\times}$ in
time~\citep{wan2025openadvancedlargescalevideo,kong2024hunyuanvideo}, some far
more~\citep{hacohen2024ltxvideorealtimevideolatent,hacohen2026ltx2efficientjointaudiovisual,chen2025deepcompressionautoencoderefficient}. A
generator can also compress through its patch
size~\citep{peebles2023scalablediffusionmodelstransformers,gupta2023photorealisticvideogenerationdiffusion}. For
images, DC-AE~\citep{chen2025deepcompressionautoencoderefficient} finds that compressing in the autoencoder beats patching. We test this for video world models and
build representation autoencoders that span seven spatio-temporal compressions.

\textbf{Few-step diffusion decoders.} Diffusion decoders avoid adversarial
losses~\citep{preechakul2022diffusionautoencodersmeaningfuldecodable,chen2025diffusionautoencodersscalableimage}
but need many steps, or a second stage to reach one~\citep{openai2023consistencydecoder,vallaeys2026ssddsinglestepdiffusiondecoder}. Shortcut
models and flow maps learn few-step generation in one stage from their own
predictions~\citep{frans2025stepdiffusionshortcutmodels,boffi2025buildconsistencymodellearning}. Our causal video decoder learns one- or two-step
decoding this way, with no adversarial loss or second stage. Appendix~\ref{app:related} extends this section.

\section{Conclusion}
\label{sec:limitations}

An interactive world model inherits its cost per frame, its action latency and much of its quality from its
autoencoder, so its compression is a design choice. To study this choice, we build
\mname, an open family spanning seven compressions, and train the same world model on each in
manipulation and navigation. \mname forms most of the latency-quality frontier, especially with a moving camera, and our study points to four findings. At strong compression, a
diffusion decoder is sharper than perceptual regression, even in one or two steps. In robot manipulation, temporal compression is more effective in the autoencoder
than in the world model. Manipulation and navigation call for opposite compressions: temporal compression costs
little when the camera is still, and spatial compression when it moves. Surprisingly, with a still camera, coarser spatial tokens hurt
motion far more than image quality, and, in robot manipulation, action following too.

Our world models each cover one environment; whether these findings carry over to a generalist one, and to policy
training and planning, remains open. Our manipulation data use still or nearly still cameras; wrist-mounted cameras move, so our account predicts they favor the compression of navigation. We hope open autoencoders will let practitioners fit a world model's compression
to its scenes and to the latency and throughput its users need.

\section*{Reproducibility statement}
Section~\ref{sec:method} and Appendix~\ref{app:ae} describe the autoencoder architecture, losses and training recipe,
and Table~\ref{tab:hparams} lists the hyperparameters of both the autoencoders and the world models. Appendix~\ref{app:data}
lists every source of the autoencoder training mixture with its sampling weight, size and preprocessing.
Appendix~\ref{app:wm} gives the world model architecture, conditioning, objective, training recipe and inference
protocol, and Appendix~\ref{app:idm} specifies the inverse dynamics probes behind action recoverability.
Section~\ref{sec:experiments} fixes the evaluation protocol shared by every autoencoder, and Table~\ref{tab:full}
reports every number behind the figures.
\ificlrfinal We release the \href{https://github.com/AdrienRR/cadre-autoencoders-and-world-models}{code} and the \href{https://huggingface.co/collections/kyutai/cadre-autoencoders-and-world-models-6abbaa26ed6191f7c975072d}{checkpoints}.\else We will release the code and the checkpoints.\fi

\section*{Ethics statement}
All training and evaluation data come from published datasets, used under their licenses. Some show people, such as
the hands in human manipulation and the faces in talking-head and action videos; we use them only to learn video
representations and do not identify anyone. Generated video can be misused to fabricate footage. Our world models are
specialized to manipulation and navigation, run at $288{\times}512$ and follow low-level actions, which limits such
use.

\section*{AI use statement}
We used generative AI tools to implement and debug code, including the training, evaluation and plotting scripts; to
draft and edit parts of the paper for readability; to create and edit figures; and to search for and format
references. We reviewed all AI-assisted work and take responsibility for the final content of this work,
including text, claims or artifacts produced with the aid of generative AI.

\bibliographystyle{iclr2027_conference}
\bibliography{refs}

\clearpage   %
\appendix
\renewcommand{\topfraction}{0.95}\renewcommand{\bottomfraction}{0.9}\renewcommand{\textfraction}{0.05}
\renewcommand{\floatpagefraction}{0.75}\setcounter{topnumber}{4}\setcounter{bottomnumber}{4}\setcounter{totalnumber}{8}

\section{Overview}
\label{app:overview}

The supplementary material details how \mname and its world models are built and scored, reports every autoencoder in
every environment and metric, and adds the analyses behind the explanations of Section~\ref{sec:compress}: long
rollouts, camera motion, counterfactual actions and seams. It also compares rollouts of every autoencoder.

\begin{tldrbox}[title={Supplementary findings at a glance}]
\begin{itemize}
\setlength{\itemsep}{2pt}\setlength{\parskip}{0pt}
\item \textbf{\mname world models run fluidly on one H100.} With 10 denoising steps, every \mname autoencoder with
  $\tau\geq4$ sustains 20 frames per second, as does \mname-$2{\times}32{\times}32$ with one decoding step, at 185~ms of
  action latency. At $\tau=1$, this needs one decoding step and at most four denoising steps
  (Appendix~\ref{app:latency}).
\item \textbf{Rollouts follow the actions, not only the usual motion of the environment.} Scored against commands
  shifted by 48 frames, which the rollout never received, every \mname autoencoder keeps a clear gap in every
  environment. The gap of Wan~2.2, Cosmos and LTX-2 shrinks to 0.04 to 0.14 in driving, where their rollouts reproduce
  forward travel (Appendix~\ref{app:idm}).
\item \textbf{Reconstruction does not predict rollout quality.} The rank correlation between reconstruction and rollout
  FVD is $+0.76$ in robot manipulation but $-0.39$ in driving and $-0.35$ indoors. Wan~2.2 reconstructs best by far, yet
  ranks fifth of twelve on rollout FVD in robot manipulation and last indoors (Appendix~\ref{app:recon}).
\item \textbf{With two decoding steps, \mname holds the best FID and FVD of every environment.} Only Cosmos ties it, on
  robot-manipulation FVD. At $2{\times}32{\times}32$, the diffusion decoder beats perceptual regression on FID and FVD in
  all four environments, and regression wins only PSNR, a sign of blur (Appendix~\ref{app:full}).
\item \textbf{Manipulation tolerates temporal compression, navigation spatial compression.} At $s=16$, robot-manipulation
  FVD stays between 39 and 46 from $\tau=1$ to 8, while from $\tau=2$ to 8 it rises by 31\% in driving and 64\% indoors.
  At $\tau=2$, $32{\times}32$ tokens multiply robot-manipulation FVD by 2.4 and lower ARR from 0.73 to 0.45 at nearly the
  same FID, but raise navigation FVD by only 11 to 12\% (Appendix~\ref{app:full}).
\item \textbf{Over 64~s, no metric collapses and \mname drifts least.} \mname-$8{\times}16{\times}16$ is best from about
  13~s on. FVD grows by 0.8 to 1.8 per second for \mname, 2.0 to 2.7 for the others and 14.5 for Wan~2.2, which starts
  best. Action recoverability barely decays, though single rollouts can still fail late (Appendix~\ref{app:drift}).
\item \textbf{The view moves far more in navigation than in manipulation.} In manipulation, content moves by about one
  pixel per frame or less, under one token over eight frames. In navigation, it moves by more than two, and half to three
  quarters of the patches change per frame instead of a third (Appendix~\ref{app:motion}).
\item \textbf{World models follow actions never recorded with their context.} From the same real context, they hold
  still or replay another clip's actions in manipulation, and turn left or right in indoor navigation
  (Appendix~\ref{app:qual}).
\item \textbf{Long latents leave seams.} At $\tau=8$, rollouts put 58 to 83\% more change at chunk boundaries than real
  video, against 8 to 47\% at $\tau=2$ (Appendix~\ref{app:seam}).
\end{itemize}
\end{tldrbox}

\begin{figure}[htbp]
\centering
\includegraphics[width=\textwidth]{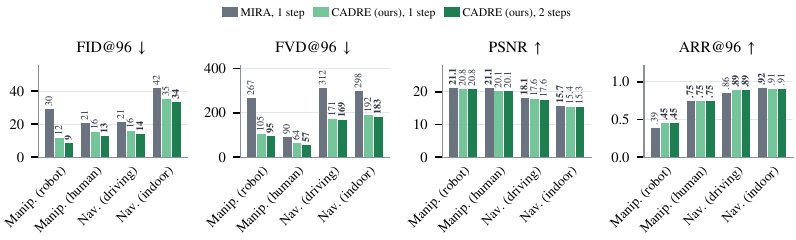}
\caption{\textbf{Diffusion against perceptual regression decoding}, all metrics (Figure~\ref{fig:decspeed}a in full), at
$2{\times}32{\times}32$. The \mname decoder lowers FID and FVD in every environment at similar action following; MIRA's
higher PSNR comes with blur.}
\label{fig:decoderfull}
\end{figure}

\textbf{Supplementary map.}
\begin{itemize}
\setlength{\itemsep}{1pt}\setlength{\parskip}{0pt}
\item \textbf{\ref{app:ae} Autoencoder implementation:} encoder layers and bottleneck strides, decoder blocks, spatial
  AdaLN, gradient balancing, losses and progressive self-distillation (Section~\ref{sec:method}).
\item \textbf{\ref{app:data} Autoencoder training data:} the 23 video sources, about 76k hours, shared by every \mname
  autoencoder and the MIRA recipe (Figure~\ref{fig:mixture}, Table~\ref{tab:mixture}).
\item \textbf{\ref{app:wm} World model:} architecture, action conditioning and training recipe
  (Table~\ref{tab:hparams}), the streaming inference loop (Figure~\ref{fig:protocol}), and the timing behind every latency
  and frame rate of the paper (Table~\ref{tab:latency}, Figure~\ref{fig:fpssteps}).
\item \textbf{\ref{app:idm} Inverse dynamics probes:} probe architecture and training, the definition of ARR, and a
  control against shifted commands (Table~\ref{tab:arrcontrol}).
\item \textbf{\ref{app:recon} Reconstruction:} held-out reconstruction of every autoencoder (Table~\ref{tab:recon}), and
  why we do not rank autoencoders by it (Section~\ref{sec:experiments}).
\item \textbf{\ref{app:full} Full results:} every autoencoder in every environment and metric, with one and two decoding
  steps (Table~\ref{tab:full}); the compression plane (Figure~\ref{fig:members}); and Figures~\ref{fig:decspeed}
  and~\ref{fig:families} in full (Figures~\ref{fig:decoderfull}, \ref{fig:speedapp}
  and~\ref{fig:familiesfull}, the last with \mname-$4{\times}24{\times}24$).
\item \textbf{\ref{app:drift} Long-horizon rollouts:} 64~s rollouts on robot manipulation (Figure~\ref{fig:drift}),
  their drift rates (Table~\ref{tab:driftrate}) and examples (Figure~\ref{fig:driftrollouts}).
\item \textbf{\ref{app:motion} Motion in the evaluation sets:} how much each environment moves (Table~\ref{tab:motion}),
  behind the camera-motion explanation of Section~\ref{sec:tauTradeoff}.
\item \textbf{\ref{app:qual} Qualitative results:} counterfactual rollouts (Figure~\ref{fig:controllability}), and rollouts of every autoencoder on the same clip in each environment (Figures~\ref{fig:uncuratedabc}
  to~\ref{fig:uncuratedre10k}).
\item \textbf{\ref{app:seam} Boundary excess:} the seam metric and its values at $\tau=2$, 4 and 8
  (Table~\ref{tab:seam}).
\item \textbf{\ref{app:related} Extended related work.}
\end{itemize}

\section{Autoencoder implementation details}
\label{app:ae}

\textbf{Encoder.} DINOv3 ViT-B/16~\citep{simeoni2025dinov3} ($p=16$) runs at $288{\times}512$, and its features average
layers 2 and 4 to 11 plus the last layer. The bottleneck convolution maps its 768 channels to 64. In time, its kernel
equals its stride $\tau$ and it has no padding, so it is causal. In space, the $18{\times}32$ grid is zero-padded to a
multiple of the stride and cropped back in the decoder. Kernel and stride are 3 for $1{\times}48{\times}48$ ($6{\times}11$
tokens), 2 for $2{\times}32{\times}32$ and 1 for $s=16$. \mname-$4{\times}24{\times}24$ uses kernel 5, stride 3 and padding 1,
and reshapes $4{\times}64$ output channels into a $2{\times}2$ block of latents ($12{\times}22$ tokens).

\textbf{Decoder.} It has 24 space-time blocks of width 1024 with 16 heads on a $16{\times}16$ patch grid; the last 4
form the temporal tail at the native frame rate. Blocks use QK-normalization, gated attention, rotary embeddings in
space and time, SwiGLU and LayerScale, with 4 spatial and 1 temporal register tokens. After the unpatchify, a refine head applies two residual blocks of dilated
depthwise and causal temporal convolutions. The decoder has about 480M parameters.

\textbf{Spatial AdaLN.}\label{app:adaln} AdaLN in DiT~\citep{peebles2023scalablediffusionmodelstransformers} modulates
all tokens of a sample with one vector. Like SPADE~\citep{park2019spade}, SEAN~\citep{zhu2020sean},
SCAM~\citep{dufour2022scam} and PixelDiT~\citep{yu2026pixelditpixeldiffusiontransformers}, spatial AdaLN modulates
each token $h_i$ from the latent $z_i$ at its position. With $c$ the embedding of the flow time $t$ and jump $d$,
\begin{equation}
(\gamma_i, \beta_i) = \mathrm{Linear}\big(\mathrm{SiLU}(z_i + c)\big),
\qquad
h_i \leftarrow (1+\gamma_i)\odot\mathrm{LN}(h_i)+\beta_i.
\label{eq:adaln}
\end{equation}
It is applied before space attention, time attention and the MLP, with separate scales and shifts. One head is
shared by all blocks, with a learned per-block offset, and is zero-initialized.

\textbf{Gradient balancing.} Each weight in
Equation~\eqref{eq:rec} is set at every step from gradient norms, as in the adaptive weight of
VQGAN~\citep{esser2021tamingtransformershighresolutionimage}:
\begin{equation}
\lambda_i \;=\; \operatorname{sg}\!\left[\frac{\|\nabla_{\mathbf{W}}\ell_{1}\|_2}{\|\nabla_{\mathbf{W}}\ell_{i}\|_2 + \delta}\right],
\label{eq:balance}
\end{equation}
where $\mathbf{W}$ is the weight of the head's last layer, $\ell_1$ the pixel term and $\ell_i$ an auxiliary term,
with $\delta=10^{-6}$ and weights clipped at $10^4$.

\textbf{Losses.} The reconstruction term of Equation~\eqref{eq:flow} is
\begin{equation}
\ell_{\text{rec}}(\hat{x}, x) \;=\; \ell_{1}(\hat{x}, x)
\;+\; \lambda_{\text{lpips}}\,\ell_{\text{lpips}}(\hat{x}, x)
\;+\; \lambda_{\text{dino}}\,\ell_{\text{dino}}(\hat{x}, x)
\;+\; \lambda_{\text{dwt}}\,\ell_{\text{dwt}}(\hat{x}, x),
\label{eq:rec}
\end{equation}
where $\ell_{\text{dino}}$ is the squared distance between normalized DINO features and $\ell_{\text{dwt}}$ the $L_1$
distance between two-level 3D Haar coefficients, which weights the high frequencies that pixel losses
underweight~\citep{degeorge2026balancing}. LPIPS and DINO use a random quarter of the frames. The wavelet term has an
extra factor of 0.25 and weights its low-pass band by 0.25.

\textbf{Progressive self-distillation.} $\mathcal{L}_{\text{psd}}$ has weight 1 on a random quarter of steps. For each
latent frame, $r$ and $t$ are two sorted uniform draws, and both half jumps use the current weights without gradients.
The reconstruction and perceptual losses supervise only $d=0$. Table~\ref{tab:hparams} gives the training recipe.

\textbf{Released autoencoders.} We also release \mbox{\mname-$2{\times}32{\times}32$}, $2{\times}16{\times}16$,
$4{\times}16{\times}16$ and $8{\times}16{\times}16$ trained beyond our experiments: to 360k steps, with the learning rate
decayed from $2{\times}10^{-4}$ to $2{\times}10^{-5}$ over the last 70k, then for 40k steps on clips of five
resolutions from $256{\times}448$ to $384{\times}640$. Our results and world models use the 150k-step versions.

\section{Autoencoder training data}
\label{app:data}

\begin{figure}[p]
\centering
\includegraphics[width=0.5\textwidth]{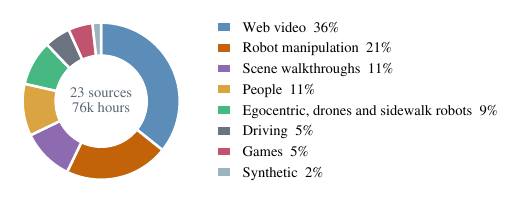}
\caption{\textbf{Autoencoder training mixture}, by share of sampling weight.}
\label{fig:mixture}
\end{figure}

Every \mname autoencoder and the MIRA-recipe baseline train on the same 23 video sources, about 76k hours
(Figure~\ref{fig:mixture}, Table~\ref{tab:mixture}). A sample is a random 16-frame window from a video of a randomly
drawn source, resized to cover $288{\times}512$, center-cropped and resampled to 20 frames per second. Videos shorter
than 16 frames or below 256 pixels on the short side are dropped.

\begin{table}[p]
\centering\small
\caption{\textbf{Sources of the autoencoder training mixture}, with their share of sampling weight and their
size.}
\label{tab:mixture}
\begin{tabular}{lp{0.52\textwidth}rr}
\toprule
Category & Source & Share (\%) & Hours \\
\midrule
Web video & Panda-70M~\citep{chen2024panda70m} & 14.3 & 10{,}330 \\
 & LiveCC~\citep{chen2025livecc} & 7.1 & 15{,}388 \\
 & LLaVA-Video-178K~\citep{zhang2024llavavideo} & 7.1 & 2{,}744 \\
 & WebVid~\citep{bain2021frozen} & 7.1 & 1{,}506 \\
\midrule
Robot manipulation & RoboCOIN, AgiBot World, Open X-Embodiment, RoboMIND, DreamDojo GR-1 and RoboCasa GR-1~\citep{wu2025robocoin,bu2025agibotworld,oneill2024openx,wu2025robomind,gao2026dreamdojo,nasiriany2024robocasa} & 8.9 & 5{,}387 \\
 & ABC, three cameras~\citep{allshire2026abc} & 5.4 & 10{,}790 \\
 & Galaxea Open-World~\citep{galaxea2025openworld} & 5.4 & 1{,}390 \\
 & HIW-500 humanoid, head camera~\citep{bitrobot2025hiw500} & 1.8 & 268 \\
\midrule
Scene walkthroughs & DL3DV-10K~\citep{ling2024dl3dv} & 7.1 & 206 \\
 & RealEstate10K~\citep{zhou2018stereo} & 3.6 & 83 \\
\midrule
People & Kinetics-700~\citep{carreira2019kinetics700} & 3.6 & 1{,}583 \\
 & CelebV-HQ and Hallo3 portraits~\citep{zhu2022celebvhq,cui2025hallo3} & 3.6 & 240 \\
 & LiveSports-3K~\citep{chen2025livecc} & 3.6 & 32 \\
\midrule
Egocentric, drones, & Sekai walking, real and game~\citep{li2025sekai} & 4.6 & 7{,}546 \\
sidewalk robots & Sekai drones, real and game~\citep{li2025sekai} & 2.9 & 109 \\
 & FrodoBots-2K~\citep{frodobots2024frodobots2k} & 1.8 & 2{,}415 \\
\midrule
Navigation (driving) & Physical AI AV, seven cameras~\citep{nvidia2025physicalaiav} & 5.4 & 6{,}019 \\
\midrule
Games & Rocket League~\citep{hu2026multiplayerinteractiveworldmodels} & 3.6 & 4{,}840 \\
 & OpenCS2~\citep{blanchon2025opencs2} & 1.4 & 5{,}084 \\
\midrule
Synthetic & Hallo3 generated portraits~\citep{cui2025hallo3} & 1.8 & 6 \\
\bottomrule
\end{tabular}
\end{table}

\section{World model architecture and training}
\label{app:wm}

\begin{figure}[t]
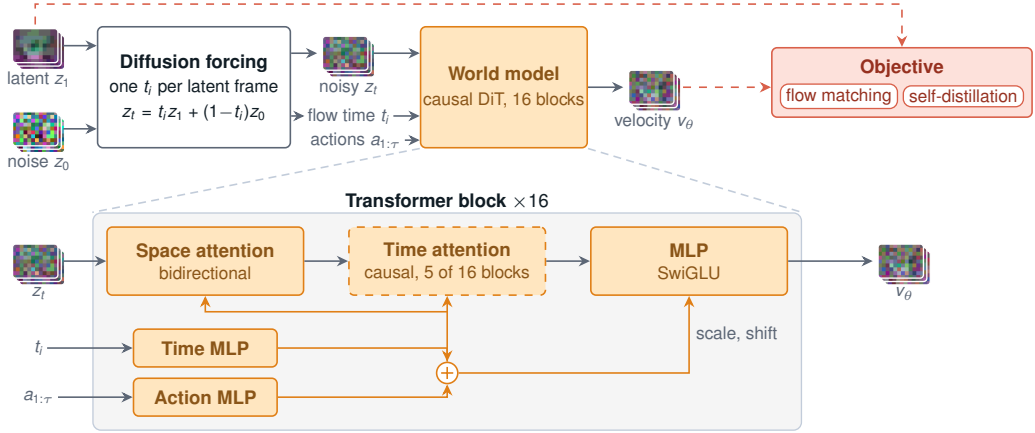

\centering
\input{figures/codec_defs}
\input{figures/wm_body}
\caption{\textbf{The world model.} Diffusion forcing gives each latent frame its own flow time; actions and flow time
condition every block through AdaLN.}
\label{fig:wm}
\end{figure}

The world model follows MIRA~\citep{hu2026multiplayerinteractiveworldmodels}, and Figure~\ref{fig:wm} is redrawn
from theirs. Table~\ref{tab:hparams} lists its hyperparameters.

\textbf{Blocks.} Each block runs bidirectional space attention within a latent frame, then a SwiGLU MLP; blocks 0, 4,
8, 12 and 15 add causal time attention in between. Attention uses 16 query and 4 key/value heads, QK-normalization, a
gate and rotary embeddings (2D in space, 1D in time). Blocks are pre-norm, without LayerScale or registers.

\textbf{Conditioning.} Each latent frame is conditioned on its flow time and on the $\tau$ actions that produced it,
action $a_j$ producing frame $j+1$. Actions are normalized per environment; their per-frame embeddings are
concatenated, projected and passed through an MLP. The flow time gets a sinusoidal embedding and an MLP, and the sum
of both embeddings drives AdaLN~\citep{peebles2023scalablediffusionmodelstransformers} before space attention, time attention and the MLP of every block. In human
manipulation, each hand contributes the frame-to-frame change in wrist position (3), wrist orientation (3) and 15
finger-joint rotations (45).

\textbf{Objective and recipe.} Under diffusion forcing~\citep{chen2024diffusionforcingnexttokenprediction}, each latent
frame draws its own flow time $t_i$, and the model regresses the velocity $z_1-z_0$ from
$z_{t_i}=t_iz_1+(1-t_i)z_0$ (Figure~\ref{fig:wm}). PSD-U, the progressive self-distillation of \citet{boffi2025buildconsistencymodellearning} with a uniformly sampled intermediate time, is added on a
fixed quarter of every batch. The recipe (Table~\ref{tab:hparams}) is the same for every autoencoder; only the action
encoder follows the environment. The world model does not patch the latent further, except in
Section~\ref{sec:placement}.

\begin{table}[htbp]
\centering\small
\caption{\textbf{Training hyperparameters} of the autoencoders and world models.}
\label{tab:hparams}
\resizebox{\textwidth}{!}{%
\begin{tabular}{lll}
\toprule
& \mname autoencoder & World model \\
\midrule
Architecture & frozen DINOv3 ViT-B/16, conv bottleneck, & causal DiT, 16 blocks, width 2048, \\
& ViT decoder 24 blocks, width 1024, 16 heads & 16 heads (4 key/value), 1.2B parameters \\
Latent & 64 channels & autoencoder latent, no further patching \\
Objective & flow matching, MAE, LPIPS, DINO, 3D Haar, PSD & flow matching, diffusion forcing, PSD-U \\
Optimizer & AdamW, $\beta=(0.9,0.95)$, no weight decay & AdamW, $\beta=(0.9,0.99)$, weight decay 0.1 \\
Learning rate & $2{\times}10^{-4}$, 2k warm-up, then constant & $10^{-4}$, 1k warm-up, then constant \\
Gradient clipping & none & 1.0 \\
Steps, batch & 150k, 96 clips (2 H100 nodes, 16 GPUs) & 100k, 64 clips (2 H100 nodes, 16 GPUs) \\
Clips & 16 frames, $288{\times}512$, 20 fps & 80 frames (40 context), $288{\times}512$, 20 fps \\
Precision, EMA & bf16, 0.9999 & bf16, 0.9999 \\
Inference & 1 or 2 decoding steps & 10 denoising steps, context noise 0.2 \\
\bottomrule
\end{tabular}}
\end{table}

\begin{figure}[t]
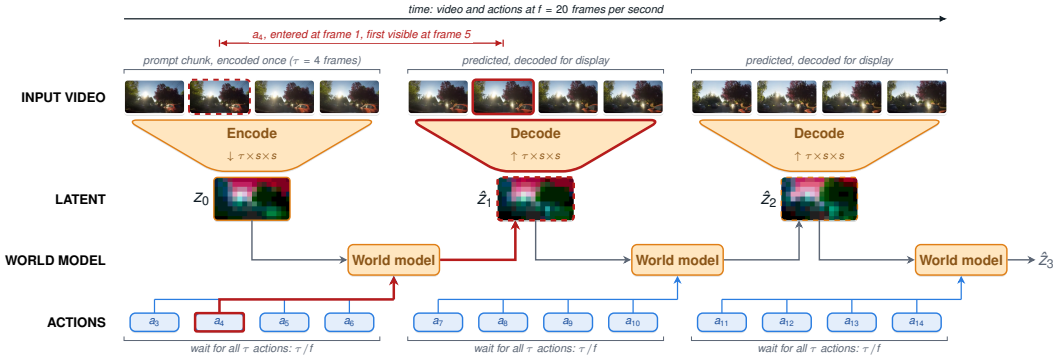

\centering
\input{figures/codec_defs}
\resizebox{\textwidth}{!}{\input{figures/action_chunk_protocol}}
\caption{\textbf{The streaming inference loop at $\tau=4$.} Each call reads the previous latent frame and the
$\tau$ actions of the chunk that just closed. Action $a_4$, entered at frame 1, first shows at frame 5 (red).}
\label{fig:protocol}
\end{figure}

\textbf{Inference.}\label{app:inference} The context is encoded once. At each chunk boundary, the world model reads the
previous latent frame and the $\tau$ actions entered during the chunk and predicts the next latent frame in 10
denoising steps. The frame is decoded for display, re-noised to level 0.2 and cached (Figure~\ref{fig:protocol}).
Since $a_t$ produces frame $t+1$, an action first shows $\tau$ frames after it is entered. This timing comes from
supervision: the architecture only keeps an action from changing frames already shown, and within a latent frame each
decoded frame can depend on all $\tau$ actions, so frame 5 reflects $a_4$ and earlier actions such as $a_3$.

\begin{figure}[htbp]
\centering
\includegraphics[width=\textwidth]{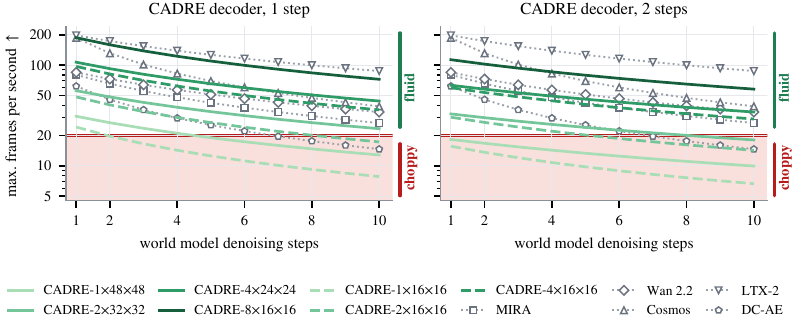}
\caption{\textbf{Can the world model run fluidly at 20 frames per second?} Maximum frame rate against denoising
steps (Table~\ref{tab:latency}); above the line, it can.}
\label{fig:fpssteps}
\end{figure}

\textbf{Latency.}\label{app:latency} Figures~\ref{fig:decspeed}b and~\ref{fig:speedapp} plot the action latency of
Equation~\eqref{eq:latency} for the first action of a chunk, the worst case, at $f=20$ frames per second. We time
$t_{\text{gen}}$, one world model call with its 10 denoising steps and cache pass, and $t_{\text{dec}}$, the decode of
one new latent frame, on one H100 at batch 1 in bf16 with \texttt{torch.compile}, averaged over 32 latent frames after
warm-up on the robot-manipulation world models (Table~\ref{tab:latency}). Each decoder runs its most efficient
streaming decode: causal decoders decode only the new latent and reuse the context (exactly for Wan~2.2; for \mname,
MIRA and Cosmos we time the decode of the new latent, the cost of a decoder cache), DC-AE decodes each frame alone,
and the non-causal LTX-2 re-decodes an 8-latent window. The rollouts of Table~\ref{tab:full} are decoded over each
decoder's trained window. The resulting maximum frame rate, $\tau/(t_{\text{gen}}+t_{\text{dec}})$, differs from the displayed rate, which is always
20 frames per second; it only tells whether the world model runs fluidly. Every \mname autoencoder with $\tau\geq4$
does, as does \mname-$2{\times}32{\times}32$ with one step; those with $\tau=1$ reach only 7 to 13 frames per second. Raising $\tau$ raises both the action latency and the maximum frame
rate (Figure~\ref{fig:teaser}). With fewer denoising steps, every
\mname autoencoder with $\tau\geq2$ runs fluidly, while at $\tau=1$ this needs one decoding step and
at most four denoising steps (Figure~\ref{fig:fpssteps}).

\begin{table}[htbp]
\centering\small
    \caption{\textbf{Timing on one H100}, each decoder at its fastest streaming decode. \mname cells: one / two decoding
steps. Max.\ fps is the sustainable rate; the display runs at 20; \textbf{bold}: fluid. $^\dagger$Non-causal, windowed.}
\label{tab:latency}
\begin{tabular}{lrrrrr}
\toprule
Autoencoder & $\tau$ & $t_{\text{gen}}$ (ms) & $t_{\text{dec}}$ (ms) & Latency (ms) & Max.\ fps \\
\midrule
\mname-$1{\times}48{\times}48$ & 1 & 56 & 22 / 44 & 128 / 150 & 12.8 / 10.0 \\
\mname-$2{\times}32{\times}32$ & 2 & 61 & 25 / 50 & 185 / 211 & \textbf{23.4} / 18.1 \\
\mname-$4{\times}24{\times}24$ & 4 & 65 & 26 / 51 & 291 / 317 & \textbf{44.0} / \textbf{34.3} \\
\mname-$8{\times}16{\times}16$ & 8 & 83 & 28 / 56 & 510 / 538 & \textbf{72.5} / \textbf{57.8} \\
\mname-$1{\times}16{\times}16$ & 1 & 106 & 22 / 45 & 178 / 200 & 7.8 / 6.7 \\
\mname-$2{\times}16{\times}16$ & 2 & 91 & 25 / 49 & 216 / 240 & 17.3 / 14.3 \\
\mname-$4{\times}16{\times}16$ & 4 & 86 & 26 / 51 & 311 / 337 & \textbf{36.0} / \textbf{29.2} \\
\midrule
MIRA recipe & 2 & 61 & 14 & 175 & \textbf{26.6} \\
Wan~2.2 & 4 & 85 & 32 & 317 & \textbf{34.2} \\
Cosmos & 8 & 198 & 7 & 605 & \textbf{39.0} \\
LTX-2$^\dagger$ & 8 & 62 & 29 & 492 & \textbf{87.4} \\
DC-AE & 1 & 64 & 5 & 118 & 14.6 \\
\bottomrule
\end{tabular}
\end{table}

\section{Inverse dynamics probes for action recoverability}
\label{app:idm}

A frozen probe, one per environment, reads the actions back from decoded frames, so it scores every autoencoder in
the same space.

\textbf{Architecture.} The probe maps a window of 8 frames to one action per frame. A frozen DINOv3
ViT-B/16~\citep{simeoni2025dinov3} gives patch features from layers 2, 5, 8 and 11, combined by a learned softmax.
Six factorized space-time blocks with bidirectional attention follow, then a per-frame attention pool and a
two-layer head (27M trained parameters).

\textbf{Training.} The probe trains on real clips from the world models' training split with a smooth $L_1$ loss on
standardized actions and photometric augmentation only. It uses AdamW (learning rate $3{\times}10^{-4}$, batch 384)
with cosine decay over 60k steps; we keep the 55k-step checkpoint (60k for human manipulation).

\textbf{Score.} For each action dimension $d$, $R^2_d$ compares the recovered actions with the commanded ones over the
last 16 generated frames of all clips, and ARR averages it normalized by the probe's $R^2$ on the real continuation,
\begin{equation}
\mathrm{ARR} = \frac{1}{|\mathcal{D}|}\sum_{d\in\mathcal{D}}\frac{\max\big(R^2_d,\,-1\big)}{R^2_d(\mathrm{real})}.
\label{eq:arr}
\end{equation}
On real video, the probe reaches $R^2$ of 0.91 in robot manipulation, 0.93 in driving and 0.83 in indoor navigation.
Human manipulation has 102 action dimensions, and the probe barely reads the 90 finger-joint dimensions at $288{\times}512$, so
$\mathcal{D}$ holds only the 12 wrist dimensions there (real-video $R^2$ 0.82).

\textbf{Control.} We also score the recovered actions against the clip's own commands shifted by 48 frames, which the
rollout never received. A world model that ignores its actions scores alike on both, so the gap between the two
raw $R^2$ values measures action following (Table~\ref{tab:arrcontrol}). Every \mname autoencoder keeps a clear gap in
every environment. In driving, the gap of Wan~2.2, Cosmos and LTX-2 is small (0.04 to 0.14),
and so is that of Cosmos and LTX-2 in human manipulation (0.24 and 0.12): their rollouts reproduce motion common to
the environment, such as forward travel in driving.

\begin{table}[t]
\centering\small
\caption{\textbf{Action recoverability and its control at frame 96.} Gap is the raw $R^2$ against the commanded
actions minus the control, against commands shifted by 48 frames. Compare gaps within an environment. \mname
rows use the two-step decode; the \mname-$1{\times}16{\times}16$ world model is trained on robot manipulation only.}
\label{tab:arrcontrol}
\setlength{\tabcolsep}{3.5pt}
\resizebox{\textwidth}{!}{%
\begin{tabular}{l ccc ccc ccc ccc}
\toprule
& \multicolumn{3}{c}{Manipulation (robot)} & \multicolumn{3}{c}{Manipulation (human)} & \multicolumn{3}{c}{Navigation (driving)} & \multicolumn{3}{c}{Navigation (indoor)} \\
\cmidrule(lr){2-4}\cmidrule(lr){5-7}\cmidrule(lr){8-10}\cmidrule(lr){11-13}
Autoencoder & ARR & control & gap & ARR & control & gap & ARR & control & gap & ARR & control & gap \\
\midrule
$1{\times}48{\times}48$ & 0.48 & $-0.99$ & 1.46 & 0.64 & $-0.54$ & 1.07 & 0.90 & 0.48 & 0.35 & 0.87 & 0.10 & 0.65 \\
$2{\times}32{\times}32$ & 0.45 & $-0.97$ & 1.40 & 0.75 & $-0.58$ & 1.19 & 0.89 & 0.47 & 0.37 & 0.91 & 0.08 & 0.70 \\
$4{\times}24{\times}24$ & 0.49 & $-0.84$ & 1.31 & 0.65 & $-0.49$ & 1.03 & 0.67 & 0.40 & 0.24 & 0.45 & 0.01 & 0.38 \\
$8{\times}16{\times}16$ & 0.72 & $-0.73$ & 1.40 & 0.63 & $-0.50$ & 1.03 & 0.86 & 0.51 & 0.29 & 0.79 & $-0.01$ & 0.69 \\
$1{\times}16{\times}16$ & 0.74 & $-0.89$ & 1.58 & N/A & N/A & N/A & N/A & N/A & N/A & N/A & N/A & N/A \\
$2{\times}16{\times}16$ & 0.73 & $-0.82$ & 1.49 & 0.73 & $-0.53$ & 1.13 & 0.97 & 0.49 & 0.41 & 0.90 & 0.05 & 0.73 \\
$4{\times}16{\times}16$ & 0.79 & $-0.77$ & 1.50 & 0.78 & $-0.53$ & 1.18 & 0.88 & 0.44 & 0.38 & 0.93 & 0.08 & 0.72 \\
\midrule
MIRA recipe & 0.39 & $-1.00$ & 1.37 & 0.75 & $-0.58$ & 1.19 & 0.86 & 0.43 & 0.38 & 0.92 & 0.03 & 0.76 \\
Wan~2.2 & 0.85 & $-0.86$ & 1.64 & 0.72 & $-0.53$ & 1.12 & 0.66 & 0.49 & 0.13 & 0.34 & 0.03 & 0.26 \\
Cosmos & 0.78 & $-0.78$ & 1.50 & $-0.27$ & $-0.44$ & 0.24 & 0.62 & 0.45 & 0.14 & 0.39 & 0.04 & 0.31 \\
LTX-2 & 0.25 & $-0.80$ & 1.05 & $-0.63$ & $-0.62$ & 0.12 & 0.57 & 0.52 & 0.04 & 0.04 & $-0.09$ & 0.15 \\
DC-AE & 0.65 & $-0.94$ & 1.55 & 0.39 & $-0.53$ & 0.85 & 0.81 & 0.33 & 0.43 & 0.86 & 0.12 & 0.62 \\
\bottomrule
\end{tabular}}
\end{table}

\section{Reconstruction}
\label{app:recon}

Table~\ref{tab:recon} reports how well each autoencoder reconstructs held-out general video. Reconstruction
does not predict rollout quality. Across autoencoders, the rank correlation between reconstruction and one-step rollout FVD
is $+0.76$ in robot manipulation but $-0.39$ in driving and $-0.35$ in indoor navigation. Wan~2.2 reconstructs best
by far, yet its world model ranks fifth of twelve on rollout FVD in robot manipulation, second to last in driving and
last in indoor navigation (Table~\ref{tab:full}).

\begin{table}[htbp]
\centering\small
\caption{\textbf{Reconstruction} of 3072 held-out OpenVid-HD~\citep{nan2025openvid1m} clips (16 frames at $288{\times}512$), with the
one-step decode for our autoencoders. No world model is involved.}
\label{tab:recon}
\begin{tabular}{llrrrr}
\toprule
& Autoencoder ($\tau{\times}s{\times}s$) & FID$\downarrow$ & FVD$\downarrow$ & PSNR$\uparrow$ & LPIPS$\downarrow$ \\
\midrule
\mname & $1{\times}48{\times}48$ & 9.71 & 43.3 & 21.21 & 0.127 \\
 & $2{\times}32{\times}32$ & 6.87 & 21.7 & 22.86 & 0.093 \\
 & $4{\times}24{\times}24$ & 7.06 & 17.7 & 23.32 & 0.087 \\
 & $8{\times}16{\times}16$ & 5.93 & 14.7 & 24.18 & 0.077 \\
 & $1{\times}16{\times}16$ & 4.27 & 10.9 & 25.05 & 0.057 \\
 & $2{\times}16{\times}16$ & 4.77 & 11.5 & 25.12 & 0.059 \\
 & $4{\times}16{\times}16$ & 4.91 & 11.2 & 24.76 & 0.065 \\
\cmidrule(l){2-6}
Others & MIRA recipe ($2{\times}32{\times}32$) & 15.73 & 127.1 & 24.44 & 0.146 \\
 & Wan~2.2 ($4{\times}16{\times}16$) & 0.78 & 0.8 & 35.02 & 0.012 \\
 & Cosmos ($8{\times}8{\times}8$) & 6.93 & 6.1 & 31.39 & 0.064 \\
 & LTX-2 ($8{\times}32{\times}32$) & 4.76 & 6.5 & 28.24 & 0.039 \\
 & DC-AE ($1{\times}32{\times}32$) & 3.59 & 22.7 & 26.83 & 0.060 \\
\bottomrule
\end{tabular}
\end{table}

\section{Full results}
\label{app:full}

\begin{figure}[htbp]
\centering
\includegraphics[width=0.5\textwidth]{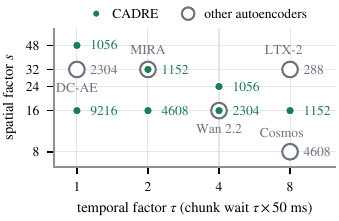}
\caption{\textbf{Compression plane.} Our seven autoencoders (green) and the five others, with the tokens per 16-frame
clip at $288{\times}512$.}
\label{fig:members}
\end{figure}

\begin{figure}[t]
\centering
\includegraphics[width=\textwidth]{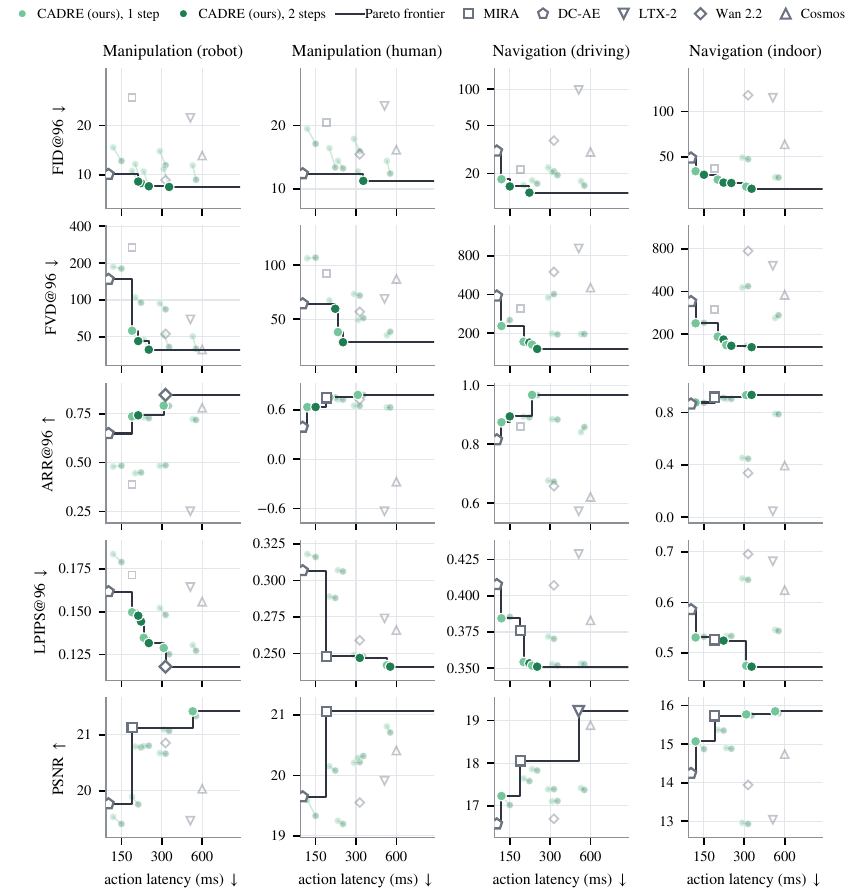}
\caption{\textbf{Rollout quality against action latency in every environment.} \mname decodes in one step (light)
or two (dark).}
\label{fig:speedapp}
\end{figure}

Table~\ref{tab:full} lists every autoencoder in every environment with both decoding step counts.
Figure~\ref{fig:members} places the autoencoders in the compression plane, and Figure~\ref{fig:speedapp} extends
Figures~\ref{fig:teaser} and~\ref{fig:decspeed}b (driving and robot manipulation) to every environment and metric.

\begin{table}[t]
\centering\small
\setlength{\tabcolsep}{4pt}
\caption{\textbf{Full results at 96 frames (4.8~s).} 2048 evaluation samples per environment, with 40 context frames
and 96 generated. Metrics use the 16 frames ending at frame 96, except indoor PSNR at frame 48, since at 4.8~s it
mostly measures viewpoint drift. 95\% confidence intervals over test episodes are about $\pm$0.2~dB PSNR ($\pm$0.4 indoors),
$\pm$0.01 LPIPS and $\pm$0.05 ARR. \mname numbers decode in two steps, with one step in parentheses. \textbf{Bold} is
best and \textit{italic} second best per environment. The $1{\times}16{\times}16$ world model is trained on robot
manipulation only, and human-manipulation ARR uses the 12 wrist dimensions, since the probe barely reads the finger joints.}
\label{tab:full}
\resizebox{\textwidth}{!}{%
\begin{tabular}{llrrrrrr}
\toprule
& Autoencoder ($\tau{\times}s{\times}s$) & Tokens & FID$\downarrow$ & FVD$\downarrow$ & PSNR$\uparrow$ & LPIPS$\downarrow$ & ARR$\uparrow$ \\
\midrule
\multicolumn{8}{l}{\textit{Manipulation (robot, ABC)}} \\
\mname & $1{\times}48{\times}48$ & 1056 & 12.15\,{\scriptsize(14.69)} & 180.3\,{\scriptsize(186.7)} & 19.41\,{\scriptsize(19.54)} & 0.179\,{\scriptsize(0.184)} & 0.48\,{\scriptsize(0.48)} \\
 & $2{\times}32{\times}32$ & 1152 & 8.92\,{\scriptsize(11.63)} & 95.2\,{\scriptsize(105.0)} & 20.78\,{\scriptsize(20.79)} & 0.144\,{\scriptsize(0.149)} & 0.45\,{\scriptsize(0.45)} \\
 & $4{\times}24{\times}24$ & 1056 & 11.46\,{\scriptsize(13.95)} & 83.9\,{\scriptsize(93.8)} & 20.66\,{\scriptsize(20.67)} & 0.148\,{\scriptsize(0.152)} & 0.49\,{\scriptsize(0.48)} \\
 & $8{\times}16{\times}16$ & 1152 & 9.34\,{\scriptsize(11.41)} & 40.1\,{\scriptsize(50.6)} & \textbf{21.33}\,{\scriptsize(21.42)} & 0.127\,{\scriptsize(0.131)} & 0.72\,{\scriptsize(0.72)} \\
 & $1{\times}16{\times}16$ & 9216 & 9.11\,{\scriptsize(10.55)} & 46.3\,{\scriptsize(56.2)} & 19.76\,{\scriptsize(19.90)} & 0.148\,{\scriptsize(0.150)} & 0.74\,{\scriptsize(0.74)} \\
 & $2{\times}16{\times}16$ & 4608 & \textit{8.53}\,{\scriptsize(10.48)} & \textbf{39.4}\,{\scriptsize(47.6)} & 20.80\,{\scriptsize(20.80)} & 0.132\,{\scriptsize(0.135)} & 0.73\,{\scriptsize(0.73)} \\
 & $4{\times}16{\times}16$ & 2304 & \textbf{8.45}\,{\scriptsize(10.86)} & 41.5\,{\scriptsize(51.9)} & 21.07\,{\scriptsize(21.09)} & \textit{0.125}\,{\scriptsize(0.129)} & \textit{0.79}\,{\scriptsize(0.79)} \\
\cmidrule(l){2-8}
Others & MIRA recipe ($2{\times}32{\times}32$) & 1152 & 29.52 & 267.1 & \textit{21.12} & 0.172 & 0.39 \\
 & Wan~2.2 ($4{\times}16{\times}16$) & 2304 & 9.29 & 52.9 & 20.85 & \textbf{0.118} & \textbf{0.85} \\
 & Cosmos ($8{\times}8{\times}8$) & 4608 & 13.10 & \textit{39.5} & 20.04 & 0.156 & 0.78 \\
 & LTX-2 ($8{\times}32{\times}32$) & 288 & 22.24 & 69.6 & 19.47 & 0.164 & 0.25 \\
 & DC-AE ($1{\times}32{\times}32$) & 2304 & 10.07 & 147.3 & 19.77 & 0.162 & 0.65 \\
\midrule
\multicolumn{8}{l}{\textit{Manipulation (human, EgoDex)}} \\
\mname & $1{\times}48{\times}48$ & 1056 & 16.38\,{\scriptsize(19.32)} & 110.1\,{\scriptsize(109.2)} & 19.33\,{\scriptsize(19.59)} & 0.316\,{\scriptsize(0.318)} & 0.64\,{\scriptsize(0.63)} \\
 & $2{\times}32{\times}32$ & 1152 & 12.66\,{\scriptsize(15.62)} & 57.2\,{\scriptsize(63.8)} & 20.08\,{\scriptsize(20.15)} & 0.288\,{\scriptsize(0.289)} & \textit{0.75}\,{\scriptsize(0.75)} \\
 & $4{\times}24{\times}24$ & 1056 & 15.10\,{\scriptsize(17.29)} & 67.7\,{\scriptsize(69.2)} & 20.22\,{\scriptsize(20.21)} & \textit{0.247}\,{\scriptsize(0.249)} & 0.65\,{\scriptsize(0.65)} \\
 & $8{\times}16{\times}16$ & 1152 & 11.84\,{\scriptsize(13.58)} & \textit{42.6}\,{\scriptsize(40.7)} & \textit{20.71}\,{\scriptsize(20.81)} & \textbf{0.241}\,{\scriptsize(0.242)} & 0.63\,{\scriptsize(0.63)} \\
 & $2{\times}16{\times}16$ & 4608 & 12.53\,{\scriptsize(13.60)} & \textbf{37.3}\,{\scriptsize(42.4)} & 19.20\,{\scriptsize(19.25)} & 0.306\,{\scriptsize(0.307)} & 0.73\,{\scriptsize(0.73)} \\
 & $4{\times}16{\times}16$ & 2304 & \textbf{10.93}\,{\scriptsize(12.15)} & 50.8\,{\scriptsize(49.5)} & 20.32\,{\scriptsize(20.29)} & 0.248\,{\scriptsize(0.248)} & \textbf{0.78}\,{\scriptsize(0.78)} \\
\cmidrule(l){2-8}
Others & MIRA recipe ($2{\times}32{\times}32$) & 1152 & 20.65 & 89.7 & \textbf{21.06} & 0.248 & 0.75 \\
 & Wan~2.2 ($4{\times}16{\times}16$) & 2304 & 14.60 & 55.0 & 19.55 & 0.259 & 0.72 \\
 & Cosmos ($8{\times}8{\times}8$) & 4608 & 15.33 & 83.8 & 20.41 & 0.266 & $-0.27$ \\
 & LTX-2 ($8{\times}32{\times}32$) & 288 & 24.78 & 64.9 & 19.91 & 0.274 & $-0.63$ \\
 & DC-AE ($1{\times}32{\times}32$) & 2304 & \textit{11.82} & 61.0 & 19.65 & 0.307 & 0.39 \\
\midrule
\multicolumn{8}{l}{\textit{Navigation (driving, Physical AI AV)}} \\
\mname & $1{\times}48{\times}48$ & 1056 & \textit{15.58}\,{\scriptsize(17.89)} & 253.1\,{\scriptsize(227.2)} & 17.02\,{\scriptsize(17.23)} & 0.385\,{\scriptsize(0.384)} & \textit{0.90}\,{\scriptsize(0.88)} \\
 & $2{\times}32{\times}32$ & 1152 & \textbf{13.84}\,{\scriptsize(15.99)} & \textit{168.7}\,{\scriptsize(171.3)} & 17.58\,{\scriptsize(17.64)} & 0.353\,{\scriptsize(0.354)} & 0.89\,{\scriptsize(0.89)} \\
 & $4{\times}24{\times}24$ & 1056 & 20.79\,{\scriptsize(22.39)} & 403.0\,{\scriptsize(378.4)} & 17.39\,{\scriptsize(17.38)} & 0.370\,{\scriptsize(0.372)} & 0.67\,{\scriptsize(0.68)} \\
 & $8{\times}16{\times}16$ & 1152 & 15.84\,{\scriptsize(17.26)} & 196.6\,{\scriptsize(196.5)} & 17.37\,{\scriptsize(17.41)} & 0.353\,{\scriptsize(0.353)} & 0.86\,{\scriptsize(0.84)} \\
 & $2{\times}16{\times}16$ & 4608 & 16.45\,{\scriptsize(17.51)} & \textbf{150.5}\,{\scriptsize(162.6)} & 17.83\,{\scriptsize(17.86)} & \textbf{0.351}\,{\scriptsize(0.352)} & \textbf{0.97}\,{\scriptsize(0.97)} \\
 & $4{\times}16{\times}16$ & 2304 & 19.34\,{\scriptsize(20.45)} & 195.3\,{\scriptsize(198.5)} & 17.11\,{\scriptsize(17.10)} & \textit{0.352}\,{\scriptsize(0.353)} & 0.88\,{\scriptsize(0.89)} \\
\cmidrule(l){2-8}
Others & MIRA recipe ($2{\times}32{\times}32$) & 1152 & 21.42 & 311.6 & 18.05 & 0.376 & 0.86 \\
 & Wan~2.2 ($4{\times}16{\times}16$) & 2304 & 37.36 & 599.7 & 16.69 & 0.407 & 0.66 \\
 & Cosmos ($8{\times}8{\times}8$) & 4608 & 30.12 & 452.5 & \textit{18.89} & 0.383 & 0.62 \\
 & LTX-2 ($8{\times}32{\times}32$) & 288 & 98.36 & 913.3 & \textbf{19.21} & 0.429 & 0.57 \\
 & DC-AE ($1{\times}32{\times}32$) & 2304 & 30.67 & 389.6 & 16.57 & 0.408 & 0.81 \\
\midrule
\multicolumn{8}{l}{\textit{Navigation (indoor, RealEstate10K); PSNR at frame 48}} \\
\mname & $1{\times}48{\times}48$ & 1056 & 38.01\,{\scriptsize(40.13)} & 240.3\,{\scriptsize(238.4)} & 14.87\,{\scriptsize(15.07)} & 0.532\,{\scriptsize(0.531)} & 0.87\,{\scriptsize(0.88)} \\
 & $2{\times}32{\times}32$ & 1152 & 33.69\,{\scriptsize(35.27)} & 182.9\,{\scriptsize(191.9)} & 15.35\,{\scriptsize(15.37)} & \textit{0.524}\,{\scriptsize(0.525)} & 0.91\,{\scriptsize(0.91)} \\
 & $4{\times}24{\times}24$ & 1056 & 48.23\,{\scriptsize(49.36)} & 435.0\,{\scriptsize(426.2)} & 12.93\,{\scriptsize(12.96)} & 0.645\,{\scriptsize(0.647)} & 0.45\,{\scriptsize(0.45)} \\
 & $8{\times}16{\times}16$ & 1152 & 36.42\,{\scriptsize(36.55)} & 271.5\,{\scriptsize(259.8)} & \textbf{15.80}\,{\scriptsize(15.85)} & 0.543\,{\scriptsize(0.546)} & 0.79\,{\scriptsize(0.79)} \\
 & $2{\times}16{\times}16$ & 4608 & \textit{33.50}\,{\scriptsize(34.13)} & \textit{165.4}\,{\scriptsize(167.3)} & 14.88\,{\scriptsize(14.90)} & 0.533\,{\scriptsize(0.533)} & 0.90\,{\scriptsize(0.90)} \\
 & $4{\times}16{\times}16$ & 2304 & \textbf{30.68}\,{\scriptsize(31.69)} & \textbf{161.7}\,{\scriptsize(165.5)} & \textit{15.74}\,{\scriptsize(15.77)} & \textbf{0.473}\,{\scriptsize(0.475)} & \textbf{0.93}\,{\scriptsize(0.93)} \\
\cmidrule(l){2-8}
Others & MIRA recipe ($2{\times}32{\times}32$) & 1152 & 41.97 & 298.5 & 15.73 & 0.526 & \textit{0.92} \\
 & Wan~2.2 ($4{\times}16{\times}16$) & 2304 & 128.35 & 771.9 & 13.94 & 0.696 & 0.34 \\
 & Cosmos ($8{\times}8{\times}8$) & 4608 & 60.53 & 377.7 & 14.74 & 0.624 & 0.39 \\
 & LTX-2 ($8{\times}32{\times}32$) & 288 & 123.61 & 606.2 & 13.04 & 0.681 & 0.04 \\
 & DC-AE ($1{\times}32{\times}32$) & 2304 & 49.13 & 339.3 & 14.24 & 0.586 & 0.86 \\
\bottomrule
\end{tabular}}
\end{table}

\begin{figure}[htbp]
\centering
\includegraphics[width=\textwidth]{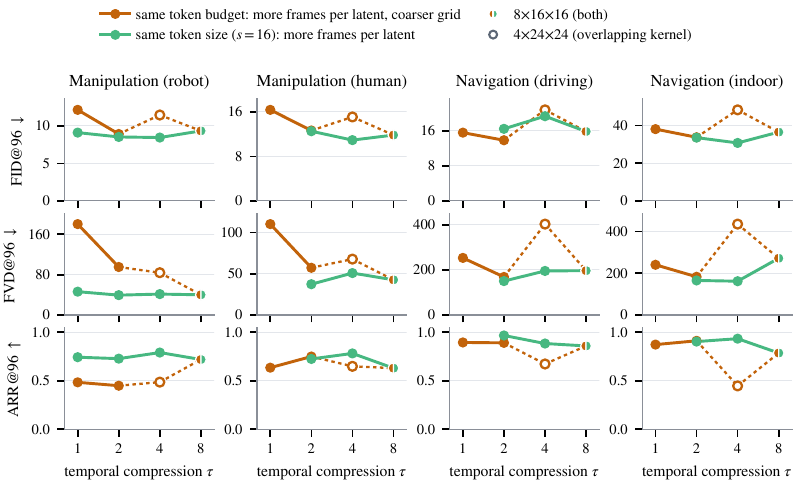}
\caption{\textbf{Temporal compression and spatial tokens, with \mname-$4{\times}24{\times}24$ in every environment.} Dotted
segments touch the overlapping-kernel cell.}
\label{fig:familiesfull}
\end{figure}

Figure~\ref{fig:familiesfull} extends Figure~\ref{fig:families} to all environments and adds \mname-$4{\times}24{\times}24$,
whose overlapping kernel confounds it; it is an outlier in navigation (FVD 403 in driving, 435 indoors). We
hypothesize that its channel-to-space reshape, which turns each position into a correlated $2{\times}2$ block of
latents, is harder to predict when the whole view moves.

Along the green curves, which vary $\tau$ at $s=16$, FVD stays between 39 and 46 in robot manipulation and ARR
between 0.72 and 0.79. In navigation, FVD rises from $\tau=2$ to $\tau=8$ by 31\% in driving, where the rise already appears at
$\tau=4$, and by 64\% in indoor navigation, where it appears only at $\tau=8$; ARR drops by 0.11 in driving and by up to 0.15 in indoor navigation. At the
same $\tau$, coarser spatial tokens cost far more in robot manipulation: at $\tau=2$, going from $s=16$ to $s=32$
multiplies FVD by 2.4 and lowers ARR from 0.73 to 0.45, with almost no change in FID, whereas in navigation FVD rises
by only 11 to 12\% and ARR changes by at most 0.08. At $\tau=1$, FVD grows more between frames 32 and 96 than at $\tau=2$ in four of
five comparisons, by 11 to 113\%, a confounded sign of error accumulation.

\section{Long-horizon rollouts}
\label{app:drift}

\begin{figure}[htbp]
\centering
\includegraphics[width=\textwidth]{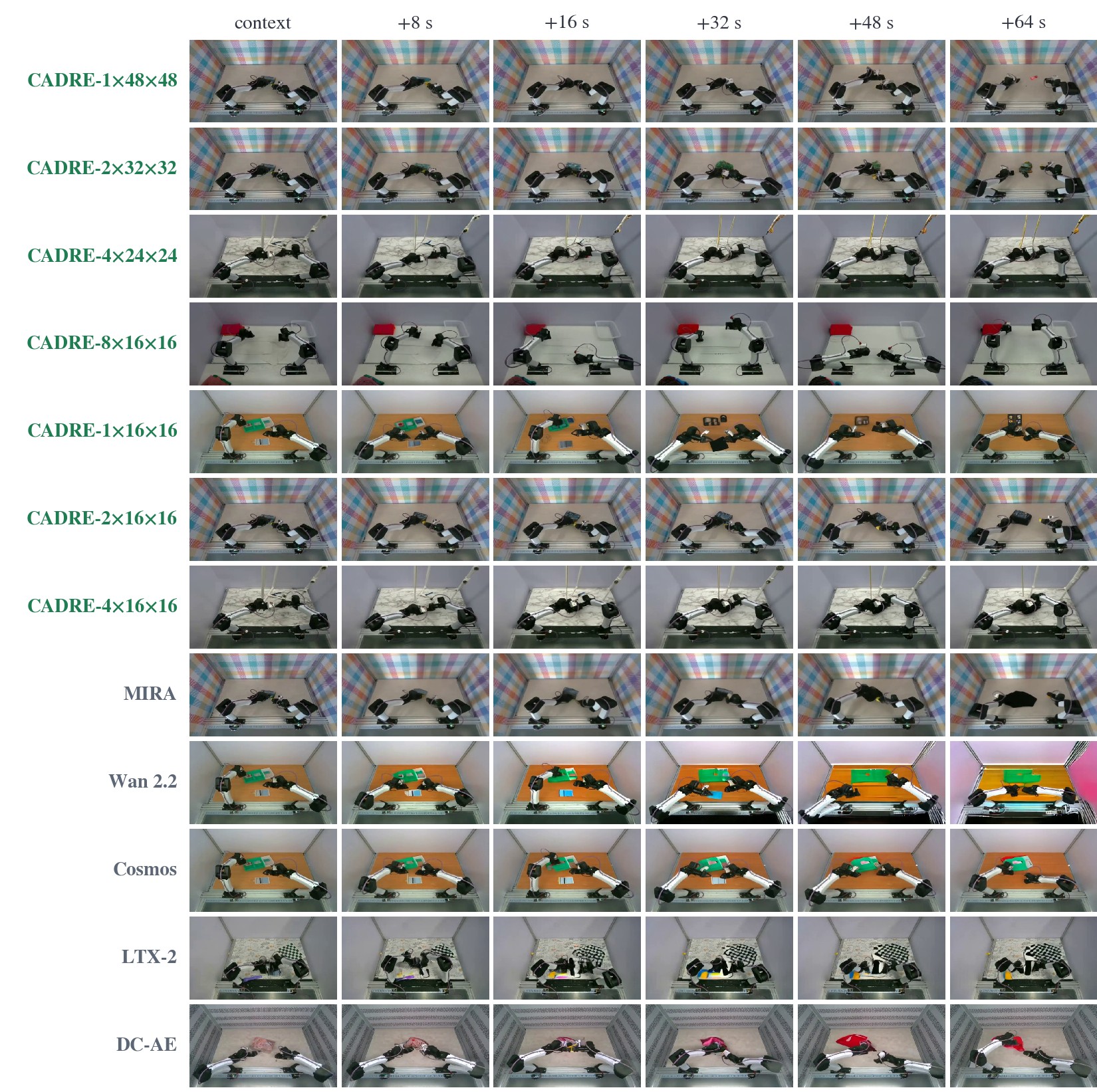}
\caption{\textbf{64~s rollouts in robot manipulation}: for each autoencoder, the real context frame, then generated frames. Scenes differ between rows.}
\label{fig:driftrollouts}
\end{figure}

\begin{figure}[htbp]
\centering
\includegraphics[width=\textwidth]{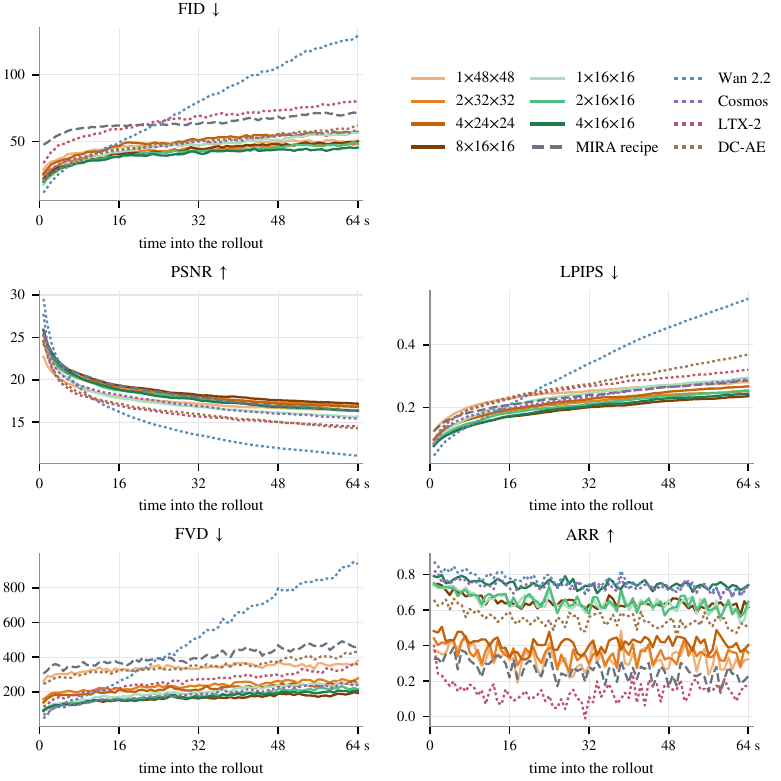}
\caption{\textbf{Rollouts over 64~s on robot manipulation.} With 256 clips, FID and FVD are biased upward; compare
them only within this figure.}
\label{fig:drift}
\end{figure}

To test whether the rankings hold beyond 4.8~s, we roll out the world models of all twelve autoencoders on robot
manipulation for 64~s (1280 frames, 256 clips, one-step decode; Figure~\ref{fig:drift}), the only environment
with many episodes long enough for such rollouts. No metric collapses.
\mname-$8{\times}16{\times}16$ is best from about 13~s on and ends with the lowest LPIPS and FVD and the highest PSNR, while the $s=16$
autoencoders lose their early lead in LPIPS and PSNR by then. Action recoverability barely decays. The other autoencoders drift faster
(Table~\ref{tab:driftrate}): FVD grows by 0.8 to 1.8 per second for the \mname autoencoders, 2.0 to 2.7 for the
others and 14.5 for Wan~2.2, which starts best but whose PSNR falls below every \mname autoencoder by about 9~s. Drift rates are
least-squares slopes between 4.8 and 64~s; the upward bias of Fr\'echet distances at 256 clips is roughly constant over
time, so slopes stay comparable. The non-causal LTX-2 decoder uses a 57-frame (8-latent) history window here.
Figure~\ref{fig:driftrollouts} shows one rollout per autoencoder; there, Wan~2.2 corrupts its colors by 48~s.

\begin{table}[htbp]
\centering\small
\caption{\textbf{Drift rates on robot manipulation}, per second between 4.8 and 64~s. Lower is better.}
\label{tab:driftrate}
\begin{tabular}{lrrrr}
\toprule
Autoencoder & FID/s & FVD/s & PSNR loss (dB/s) & LPIPS/s ($\times10^{-3}$) \\
\midrule
$1{\times}48{\times}48$ & \textbf{0.15} & \textbf{0.8} & \textbf{0.046} & \textbf{1.4} \\
$2{\times}32{\times}32$ & 0.19 & 1.2 & 0.060 & 1.6 \\
$4{\times}24{\times}24$ & 0.29 & 1.1 & 0.058 & 1.8 \\
$8{\times}16{\times}16$ & 0.25 & 0.9 & 0.059 & 1.6 \\
$1{\times}16{\times}16$ & 0.35 & 1.8 & 0.062 & 2.1 \\
$2{\times}16{\times}16$ & 0.24 & 1.5 & 0.065 & 1.8 \\
$4{\times}16{\times}16$ & 0.19 & 1.3 & 0.068 & 1.8 \\
\midrule
MIRA recipe & 0.22 & 2.4 & 0.077 & 1.8 \\
Wan~2.2 & 1.75 & 14.5 & 0.140 & 7.4 \\
Cosmos & 0.35 & 2.0 & 0.071 & 2.1 \\
LTX-2 & 0.47 & 2.7 & 0.066 & 2.2 \\
DC-AE & 0.41 & 2.1 & 0.073 & 3.0 \\
\bottomrule
\end{tabular}
\end{table}

\section{Motion in the evaluation sets}
\label{app:motion}

Table~\ref{tab:motion} measures motion on the real continuation of every evaluation clip: the mean frame-to-frame
pixel difference, the share of pixels and of $32{\times}32$ patches that change by more than 0.05, and the
RAFT~\citep{teed2020raft} optical flow in pixels per frame.

\begin{table}[htbp]
\centering\small
\caption{\textbf{How much each environment moves.} Averages over the 2048 evaluation samples of each environment
(1706 distinct clips indoors).}
\label{tab:motion}
\begin{tabular}{lcccc}
\toprule
& Pixel difference & Pixels that change & Patches that change & Optical flow \\
Environment & per frame & per frame & per frame & (pixels per frame) \\
\midrule
Manipulation (robot) & 0.010 & 2.6\% & 32\% & 0.24 \\
Manipulation (human) & 0.009 & 3.6\% & 33\% & 1.27 \\
Navigation (driving) & 0.017 & 8.2\% & 48\% & 2.13 \\
Navigation (indoor) & 0.032 & 15.7\% & 76\% & 2.69 \\
\bottomrule
\end{tabular}
\end{table}

In manipulation, a third of the image changes per frame and content moves by about one pixel per frame or less, less
than one 16-pixel token over eight frames. In navigation, half to three quarters of the image changes and content
moves by more than two pixels per frame, more than a token over eight frames. The head-mounted camera of human
manipulation shifts the whole image slightly, which optical flow captures and pixel differences miss.

\section{Qualitative results}
\label{app:qual}

\begin{figure}[p]
\centering
\includegraphics[width=\textwidth]{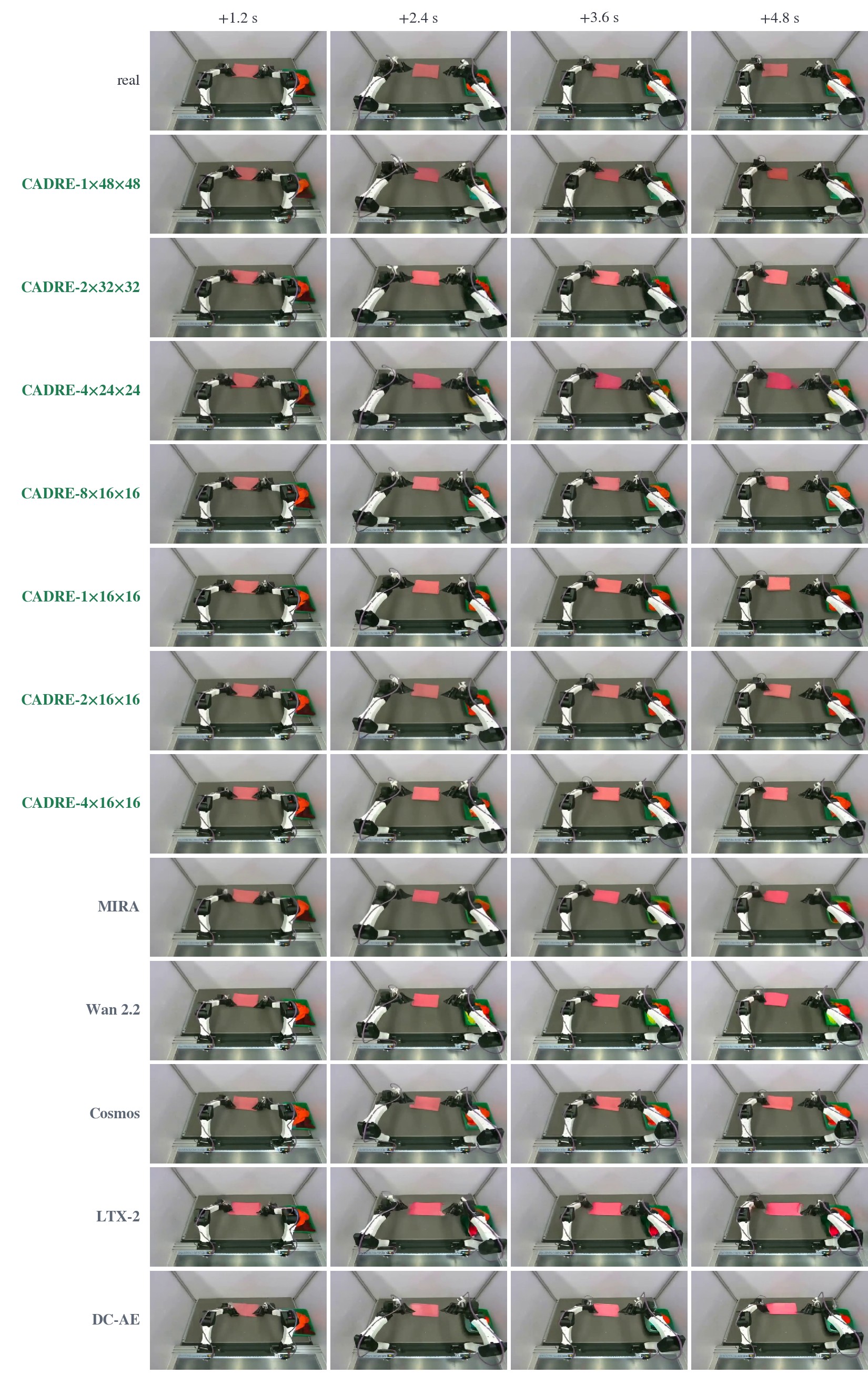}
\caption{\textbf{Rollouts, robot manipulation}: the same clip and recorded actions for every autoencoder, with the real future on top.}
\label{fig:uncuratedabc}
\end{figure}

\begin{figure}[p]
\centering
\includegraphics[width=\textwidth]{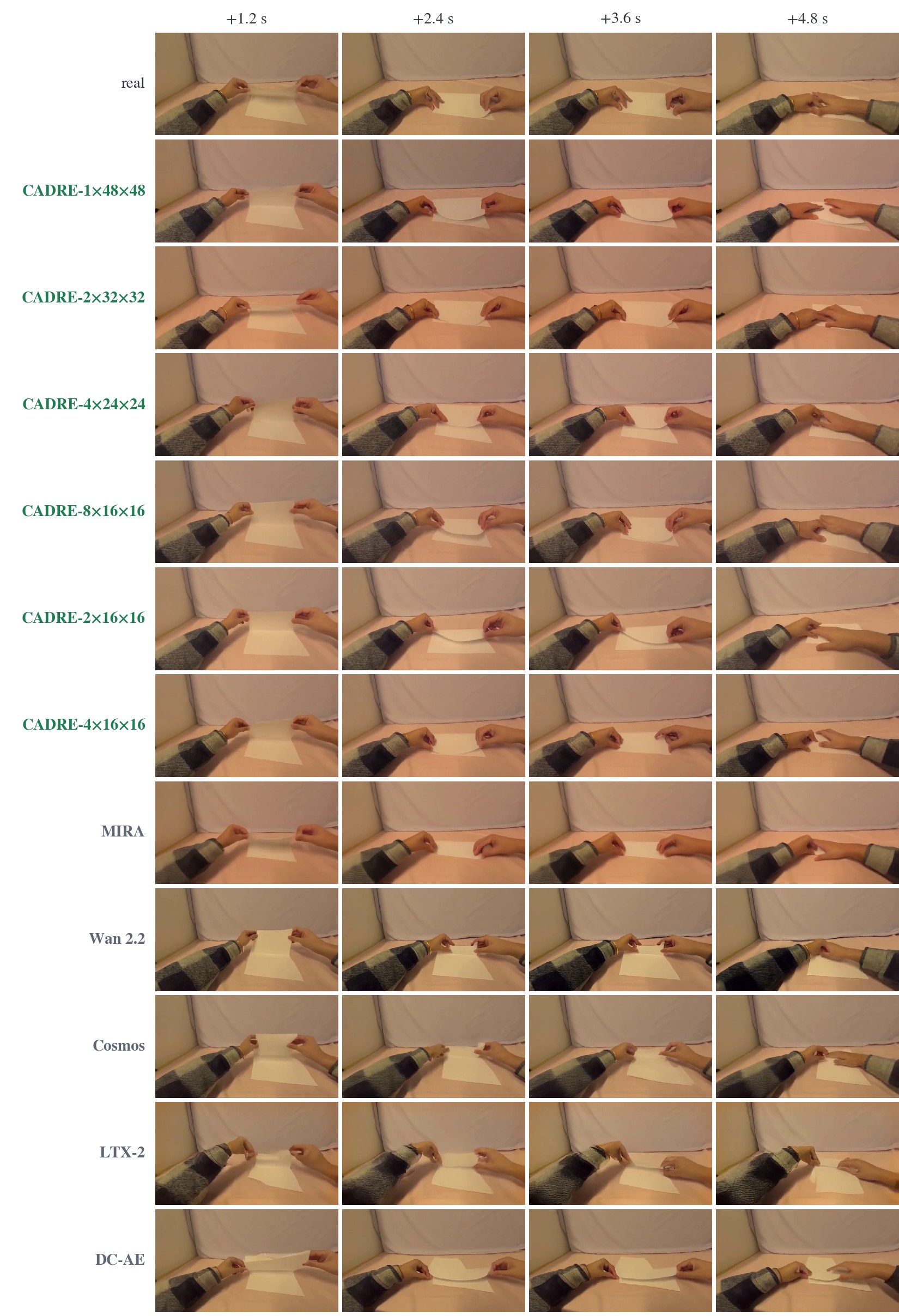}
\caption{\textbf{Rollouts, human manipulation}: the same clip and recorded actions for every autoencoder, with the real future on top.}
\label{fig:uncuratedegodex}
\end{figure}

\begin{figure}[p]
\centering
\includegraphics[width=\textwidth]{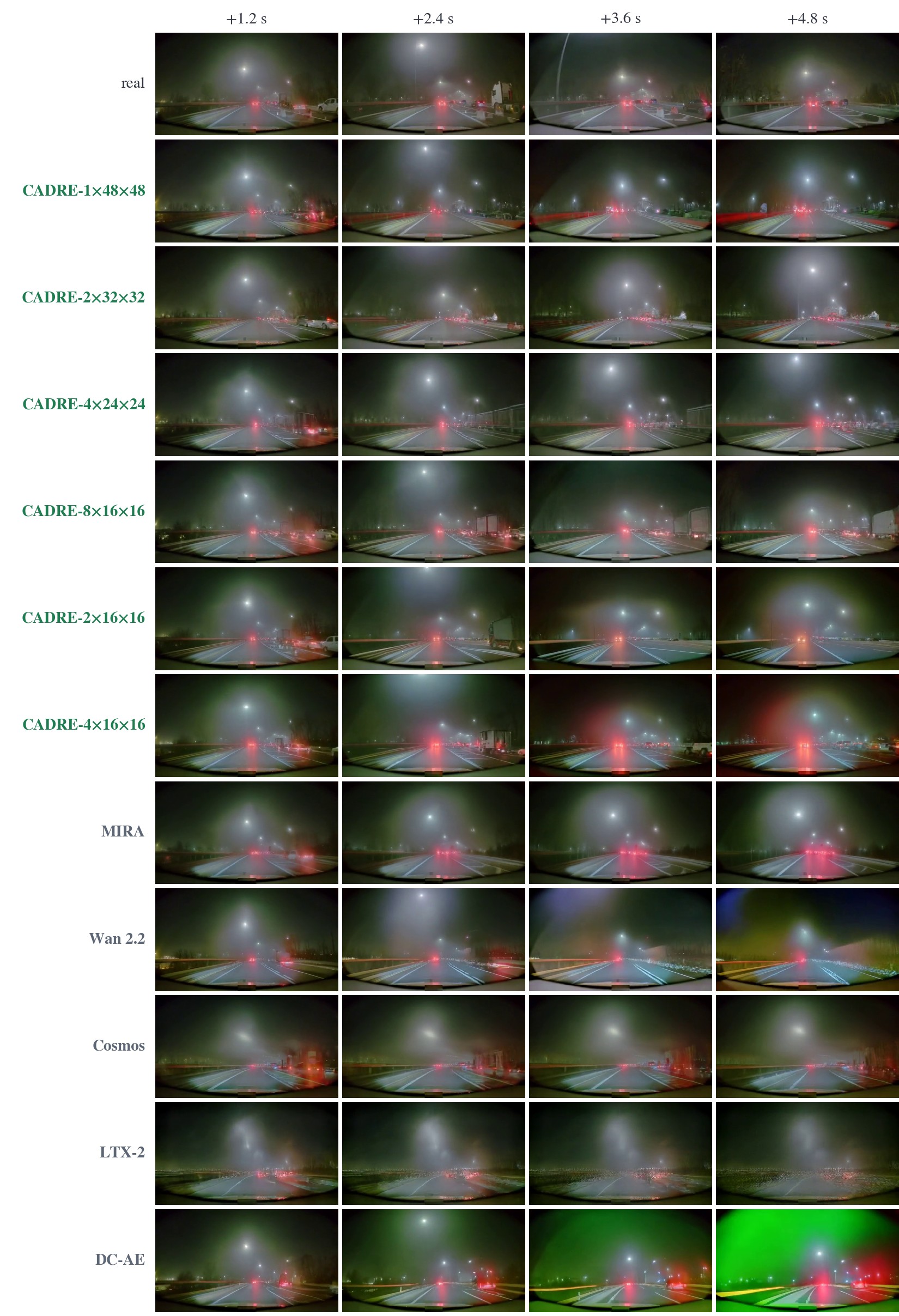}
\caption{\textbf{Rollouts, driving}: the same clip and recorded actions for every autoencoder, with the real future on top.}
\label{fig:uncurateddriving}
\end{figure}

\begin{figure}[p]
\centering
\includegraphics[width=\textwidth]{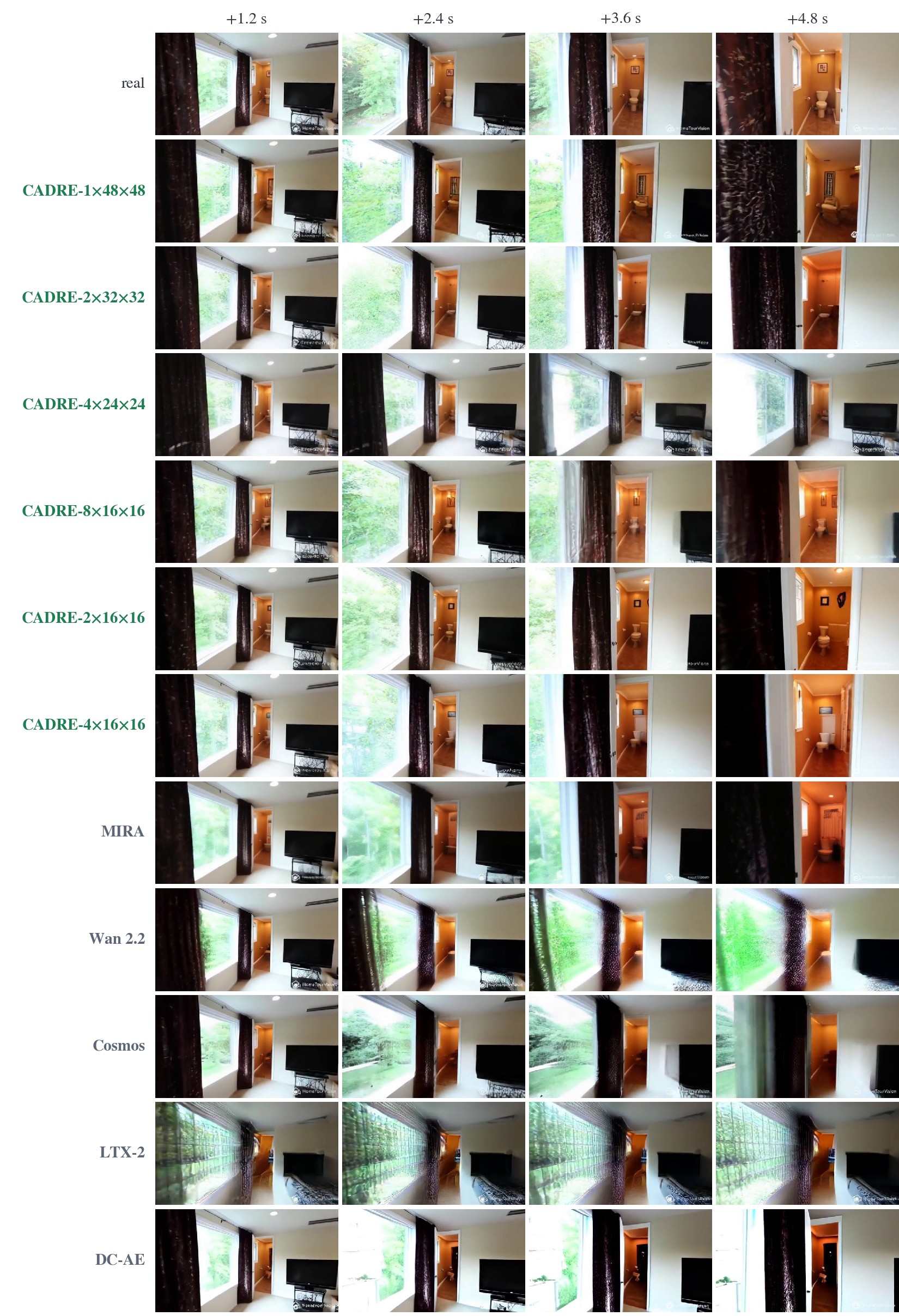}
\caption{\textbf{Rollouts, indoor navigation}: the same clip and recorded actions for every autoencoder, with the real future on top.}
\label{fig:uncuratedre10k}
\end{figure}

\begin{figure}[htbp]
\centering
\includegraphics[width=\textwidth]{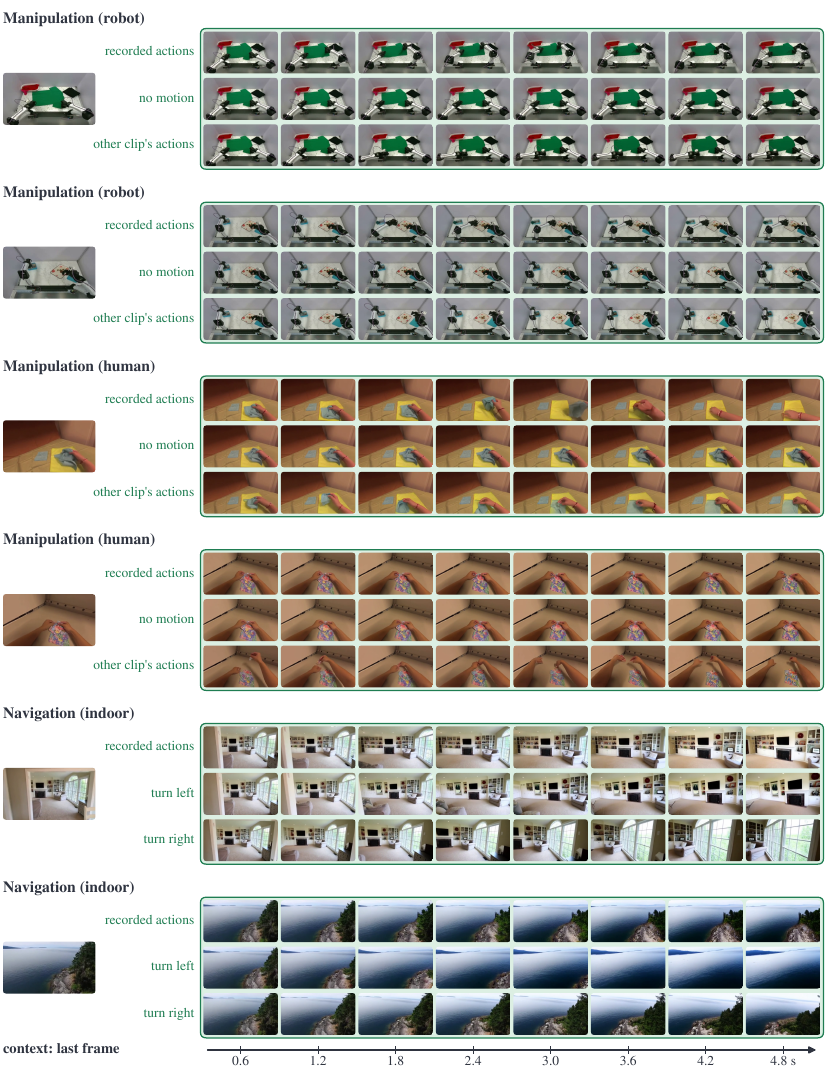}
\caption{\textbf{Counterfactual rollouts.} From one real context (left), the world model follows the recorded actions
and two counterfactual ones (green): no motion or another clip's actions in manipulation, a constant left or right
turn in indoor navigation.}
\label{fig:controllability}
\end{figure}

Figure~\ref{fig:controllability} shows counterfactual rollouts: from the same real context, the world model follows the
recorded actions and two action sequences never recorded with it. Rollouts under recorded actions are in
Figures~\ref{fig:teaser} and~\ref{fig:qualitative}.
Figures~\ref{fig:uncuratedabc} to~\ref{fig:uncuratedre10k} compare every autoencoder on the same clip in each
environment, with its recorded actions.

\section{Boundary excess}
\label{app:seam}

Let $d_t$ be the mean absolute difference between generated frames $t$ and $t-1$, with pixels in $[0,1]$. $E$ is the
ratio of the mean $d_t$ over boundary pairs ($t$ a multiple of $\tau$) to that over the other pairs, and the boundary
excess is $E/E_{\text{real}}$, with $E_{\text{real}}$ from the real continuation at the same indices. One means no
seam. We compute it on 32 clips per autoencoder with the one-step decode; repeated draws differ by about 0.15. At
$\tau=8$, rollouts put 58 to 83\% more change at chunk boundaries than real video (Table~\ref{tab:seam}).

\begin{table}[htbp]
\centering\small
\caption{\textbf{Boundary excess} at $\tau=2$, 4 and 8 ($2{\times}16{\times}16$, $4{\times}16{\times}16$,
$8{\times}16{\times}16$; $2{\times}32{\times}32$, $4{\times}24{\times}24$ and $8{\times}16{\times}16$ in human manipulation). One
means no seam.}
\label{tab:seam}
\begin{tabular}{lccc}
\toprule
Environment & $\tau=2$ & $\tau=4$ & $\tau=8$ \\
\midrule
Manipulation (robot) & 1.17 & 1.34 & 1.81 \\
Manipulation (human) & 1.47 & 1.56 & 1.75 \\
Navigation (driving) & 1.22 & 1.34 & 1.83 \\
Navigation (indoor) & 1.08 & 1.18 & 1.58 \\
\bottomrule
\end{tabular}
\end{table}

\FloatBarrier
\section{Extended related work}
\label{app:related}

We study how much the autoencoder of an interactive world model compresses space and time, and whether
that compression belongs in the autoencoder or in the world model.

\textbf{World models, from offline to streaming.} A world model predicts how an environment evolves under actions. Early ones predict compact latents for
control~\citep{ha2018worldmodels,hafner2020dreamer,micheli2023iris}; most recent ones generate the video
itself~\citep{bai2023sequentialmodelingenablesscalable}, for
driving~\citep{gao2024vistageneralizabledrivingworld,russell2025gaia2controllablemultiviewgenerative,carrion2025tardisstridespatiotemporalroad},
robotics~\citep{yang2024unisim}, egocentric motion~\citep{bai2025wholebodyconditionedegocentricvideo,wang2026liftingembodiedworldmodels}
and open scenes~\citep{bruce2024genie,mao2025yumeinteractiveworldgeneration,panteam2026panworldmodelgeneral,zhu2026astrageneralinteractiveworld}.
Most are offline, taking seconds to minutes per clip. They compare autoencoders only in that regime:
SANA-WM~\citep{zhu2026sanawmefficientminutescaleworld} swaps the Wan~2.1 autoencoder for LTX-2's. An interactive world model streams instead and must render each frame before the next action
arrives. Some reach real time by rendering directly from
controls~\citep{kim2020learningsimulatedynamicenvironments,menapace2021playablevideogeneration} or with
dedicated models~\citep{valevski2024gamengen,alonso2024diamond,bar2025navigationworldmodels,rahary2026viewenoughmonoculartraining}.
Others adapt a large video generator by streaming
distillation~\citep{yin2025slowbidirectionalfastautoregressive,huang2025selfforcingbridgingtraintest}, as
Matrix-Game 2.0~\citep{he2025matrixgame2} and Hunyuan-GameCraft~\citep{li2025hunyuangamecraft} do. They inherit its autoencoder unchanged, at $8{\times}$ in space and $4{\times}$ in time. Their blockwise generation
also adds serial computation~\citep{chao2026serialitygapvideodiffusion}.
MineWorld~\citep{guo2025mineworldrealtimeopensourceinteractive} uses discrete tokens instead. Once
generation is fast, the decoder becomes the bottleneck, which Matrix-Game
3.0~\citep{wang2026matrixgame30realtimestreaming} and WorldPlay~\citep{sun2026worldplaylongtermgeometricconsistency}
prune. MIRA~\citep{hu2026multiplayerinteractiveworldmodels} instead trains its own autoencoder and judges
it by the world model; we extend that principle along both compression axes.

\textbf{Diffusion and flow-based generation.} Both the world model and the decoder of \mname are diffusion models trained with flow matching. Diffusion
models~\citep{sohldickstein2015deepunsupervisedlearning,ho2020denoisingdiffusionprobabilistic,song2019generativemodelingestimating,song2020scorebasedgenerativemodeling}
and their refinements of sampler, schedule, loss weighting and
preconditioning~\citep{song2020denoisingdiffusionimplicit,lu2022dpmsolverfastode,nichol2021improveddenoisingdiffusion,kingma2021variationaldiffusionmodels,kingma2023understandingdiffusionobjectives,hang2023efficientdiffusiontraining,karras2022elucidatingdesignspace,karras2023analyzingimprovingtraining}
converge in flow matching~\citep{lipman2022flowmatchinggenerative,liu2022flowstraightfast,albergo2023stochasticinterpolantsunifying},
which regresses a velocity along an interpolant between noise and data and scales with
transformers~\citep{bao2022allworthwords,peebles2023scalablediffusionmodelstransformers,ma2024sitexploringflow,esser2024scalingrectifiedflowtransformers}.
Three of its design choices interact with compression. The first is the prediction target. Noise and velocity
prediction~\citep{salimans2022progressivedistillationfastsampling} degrade when each token is high-dimensional,
and clean-sample prediction does not~\citep{li2026basicsletdenoisinggenerative}. Our decoder outputs large pixel
patches, so it predicts the clean sample, and its wavelet loss, related to the frequency
balancing of \citet{degeorge2026balancing}, emphasizes the fine texture it must restore. The world model's tokens keep
64 channels at every compression, so it keeps velocity prediction. The second is the noise schedule, whose optimum
shifts with the number of tokens~\citep{chen2023importancenoisescheduling,hoogeboom2025simplerdiffusionsid215};
RAE~\citep{zheng2025diffusiontransformersrepresentationautoencoders} generalizes the shift of
\citet{esser2024scalingrectifiedflowtransformers} to the number of tokens times their channels. Our compressions
change the number of tokens by more than an order of magnitude, so our tuning shifts the schedule with each autoencoder's latent size and token count. The third is the representation. Aligning a generator's features to a self-supervised encoder
speeds up its training~\citep{yu2024representationalignmentgeneration}, as masking
does~\citep{gao2023mdtv2maskeddiffusion}; we place that prior in the autoencoder instead, and the world model
generates in the features of a frozen DINOv3.

\textbf{Conditioning.} The world model is conditioned on actions, and how faithfully it follows them is one of our three axes. Image
generators usually strengthen adherence to their condition with
guidance~\citep{dhariwal2021diffusionmodelsbeat,ho2022classifierfreediffusionguidance,kynkaanniemi2024applyingguidancelimited,karras2024guidingdiffusionmodel};
we use none, so action following reflects the latent and the model alone. A condition can also be weak, as with noisy
web captions~\citep{dai2023emuenhancingimage,podell2023sdxlimprovinglatent}, and some action dimensions barely change
the image, such as the finger joints in human manipulation, which our probe cannot read even on real video. We
therefore measure action following directly, by how well an inverse dynamics probe recovers the actions from the
generated frames (Appendix~\ref{app:idm}).

\textbf{Evaluation.} We judge each autoencoder by the world model it yields, with FID~\citep{heusel2018ganstrainedtimescaleupdate} and
FVD~\citep{unterthiner2019accurategenerativemodelsvideo}. These metrics are fragile. FID mixes fidelity and
diversity~\citep{kynkaanniemi2019improvedprecisionrecall}, depends on resizing and ImageNet
features~\citep{parmar2021aliasedresizingsurprising,kynkaanniemi2022roleimagenetclasses}, can disagree with human
judgment~\citep{stein2023exposingflawsgenerative,jayasumana2023rethinkingfidtowards}, and moves more when the
model is retrained with another seed than when it is resampled~\citep{dufour2026fidlotteryquantifying}.

\textbf{Video and causal generation.} Video diffusion moved from pixel space~\citep{ho2022videodiffusionmodels} to latents, first by inflating image
networks~\citep{blattmann2023stablevideodiffusion} and then with spatio-temporal
transformers~\citep{yang2025cogvideoxtexttovideodiffusionmodels,kong2024hunyuanvideo,polyak2024moviegencast,wan2025openadvancedlargescalevideo},
which denoise a whole clip at one noise level. Causal generation breaks that symmetry. Rolling
diffusion~\citep{ruhe2024rollingdiffusionmodels} gives later frames more noise, and diffusion
forcing~\citep{chen2024diffusionforcingnexttokenprediction} draws one noise level per frame, so the next frame is
denoised from a clean or partly noisy past. History guidance~\citep{song2025historyguidedvideodiffusion} builds
on it, and MAGI-1~\citep{teng2025magi1autoregressivevideo} scales chunk-wise autoregression to 24B parameters.
Noise on the context~\citep{valevski2024gamengen} and training on the model's own
rollouts~\citep{huang2025selfforcingbridgingtraintest} curb drift over long horizons, and
FramePack~\citep{zhang2025framecontextpacking} compresses the history to bound its cost. Discrete alternatives
predict quantized tokens
autoregressively~\citep{yu2024languagemodelbeatsdiffusion,bruce2024genie,wu2024ivideogptinteractivevideogpts} or
by masked prediction~\citep{chang2022maskgitmaskedgenerative}, and MAR~\citep{li2024autoregressiveimagegeneration}
keeps autoregression but replaces the quantizer with a per-token diffusion loss. Our world model combines flow
matching, diffusion forcing and context noise, and stays fixed while the autoencoder varies.

\textbf{The autoencoder as an object of study.} In latent diffusion~\citep{rombach2022highresolutionimagesynthesislatent,esser2021tamingtransformershighresolutionimage},
generation quality depends on the structure of the latent. A line of work improves it at a fixed generator
(Section~\ref{sec:experiments}).
Representation autoencoders go further and generate in the space of a frozen self-supervised
encoder~\citep{zheng2025diffusiontransformersrepresentationautoencoders,singh2026improvedbaselinesrepresentationautoencoders},
for text-to-image~\citep{tong2026scalingtexttoimagediffusiontransformers}, low-dimensional
latents~\citep{gao2025layerenoughadaptingpretrained,shi2026latentdiffusionmodelvariational,bi2026vfmvaevisionfoundationmodels}
and video~\citep{xie2026videoraetamingvideofoundation,guo2026vraerethinkingvideolatent,hu2026multiplayerinteractiveworldmodels},
on encoders such as
DINO~\citep{caron2021emergingpropertiesselfsupervisedvision,oquab2024dinov2learningrobustvisual,simeoni2025dinov3}
and V-JEPA~2~\citep{assran2025vjepa2selfsupervisedvideo,murlabadia2026vjepa21unlockingdense}. Distilling
such an encoder into the latent, as in GAIA~\citep{hu2023gaia1generativeworldmodel,russell2025gaia2controllablemultiviewgenerative}
and Mimi~\citep{defossez2024moshispeechtextfoundationmodel}, follows the same idea and makes
latents easier to diffuse. These works use one compression rate; our representation autoencoders vary it in space and time,
one trained autoencoder per compression point.

\textbf{Video autoencoders and strong compression.} Video autoencoders use causal 3D convolutions~\citep{yu2024languagemodelbeatsdiffusion}. Most settle
at $8{\times}$ in space and $4{\times}$ in
time~\citep{yang2025cogvideoxtexttovideodiffusionmodels,wan2025openadvancedlargescalevideo,kong2024hunyuanvideo}.
Others compress harder: LTX-Video~\citep{hacohen2024ltxvideorealtimevideolatent},
DC-AE~\citep{chen2025deepcompressionautoencoderefficient},
DC-VideoGen~\citep{chen2025dcvideogenefficientvideogeneration}, H3AE~\citep{wu2025h3aehighcompressionhigh},
ProMAG~\citep{mahapatra2025progressivegrowingvideotokenizers} and
DeltaTok~\citep{kerssies2026frameworthtokenefficient}, down to one token per frame. Comparisons across
rates use reconstruction~\citep{tang2024vidtokversatileopensourcevideo,cosmos2025worldfoundationmodelplatform}
or text-to-video generation~\citep{hansenestruch2025learningsscalingvisualtokenizers}. The two rates
usually move together. We vary them independently and judge each by an action-conditioned world model.

\textbf{Diffusion decoders and single-step generation.} Pixel and perceptual losses alone blur, so autoencoders add an adversarial
loss~\citep{esser2021tamingtransformershighresolutionimage}, at the price of a sensitive discriminator.
Diffusion decoders are the alternative~\citep{preechakul2022diffusionautoencodersmeaningfuldecodable,pandey2022diffusevaeefficientcontrollablehighfidelity,pernias2023wuerstchenefficientarchitecturelargescale},
and scale as tokenizers with a single flow-matching loss~\citep{chen2025diffusionautoencodersscalableimage},
but decode in many steps. Single-step diffusion decoders so far need a second stage or an adversarial
term~\citep{vallaeys2026ssddsinglestepdiffusiondecoder,sargent2025flowmodemodeseekingdiffusion,zhao2025epsilonvaedenoisingvisualdecoding};
video diffusion decoders keep the adversarial loss~\citep{hacohen2024ltxvideorealtimevideolatent} or several
steps~\citep{yang2025rethinkingvideotokenizationconditioned,teng2026adaptive1dvideodiffusion}.
Few-step generation itself comes from distillation~\citep{salimans2022progressivedistillationfastsampling,luo2023latentconsistencymodelssynthesizing,sauer2023adversarialdiffusiondistillation}
and consistency models~\citep{song2023consistencymodels,lu2025simplifyingstabilizingscalingcontinuoustime};
shortcut models~\citep{frans2025stepdiffusionshortcutmodels}, flow maps~\citep{boffi2025buildconsistencymodellearning}
and MeanFlow~\citep{geng2025meanflowsonestepgenerative} learn it in one stage from their own predictions.
We bring that one-stage self-distillation to a causal decoder with no discriminator.

\textbf{Where compression lives.} A generator can also compress through its patch size.
DiT~\citep{peebles2023scalablediffusionmodelstransformers} improves as the patch shrinks, pixel-space
transformers~\citep{li2026basicsletdenoisinggenerative,hoogeboom2025simplerdiffusionsid215} replace the
autoencoder with a large patch, W.A.L.T.~\citep{gupta2023photorealisticvideogenerationdiffusion} and
Latte~\citep{ma2025lattelatentdiffusiontransformer} find temporal patching costliest, and
FlexiDiT~\citep{anagnostidis2025flexiditdiffusiontransformereasily} varies the patch per step. For images,
DC-AE~\citep{chen2025deepcompressionautoencoderefficient} finds that compressing in the autoencoder beats
patching at matched tokens. We make that comparison for a video world model.

\ificlrfinal   %
\section*{Acknowledgments}
We thank Hedi Ben-Younes, Moritz B\"ohle, Andreas Geiger, Alyx Liao, Renaud Marlet, Juliette Marrie, Manu Orsini, Am\'elie Royer,
Nicolas Sereyjol-Garros and V\'aclav Volhejn for interesting discussions.
\fi
\end{document}